\documentclass[preprint,authoryear,11pt]{elsarticle}
\usepackage{geometry}
\usepackage{multirow}

\usepackage{graphicx}
\usepackage{subcaption}
\usepackage{fancyhdr}
\usepackage{amsmath}
\usepackage{xcolor}
\usepackage{amssymb}
\usepackage{amsthm}
\usepackage{multirow}
\usepackage{float}
\usepackage{marvosym}
\usepackage{natbib}
\usepackage{bm}

\usepackage{algorithm}
\usepackage{algorithmic}

\usepackage{multirow}
\usepackage{algorithm}
\usepackage{algorithmic}
\usepackage{textcomp}
\usepackage{marvosym}
\usepackage{longtable}

\newtheorem{remark}{Remark}
\newtheorem{example}{Example}

\biboptions{sort&compress}

\begin{document}

\begin{frontmatter}

\title{BooleanOCT: Optimal Classification  Trees  based on multivariate Boolean Rules}

\author{Jiancheng Tu$^1$, \quad Wenqi Fan$^{2,3}$, \quad Zhibin Wu$^1$}
\address{1 Business School, Sichuan University, Chengdu 610065, China\\}

\address{2 Department of Computing, The Hong Kong Polytechnic University}
\address{{3 Department of Management and Marketing, The Hong Kong Polytechnic University}}

\begin{abstract}
{The global optimization of classification trees has demonstrated considerable promise, notably in enhancing accuracy, optimizing size, and thereby improving human comprehensibility.
While existing optimal classification trees substantially enhance accuracy over greedy-based tree models like CART, they still fall short when compared to the more complex black-box models, such as random forests.
To bridge this gap, we introduce a new mixed-integer programming (MIP) formulation, grounded in multivariate  Boolean rules, to derive the optimal classification tree.
Our methodology  integrates both linear metrics, including accuracy, balanced accuracy, and cost-sensitive cost, as well as nonlinear metrics such as the F1-score.
The approach is implemented in an open-source Python package named \textit{BooleanOCT}.
We comprehensively benchmark these methods on the 36 datasets from the UCI machine learning repository.
The proposed models demonstrate practical solvability on real-world datasets, effectively handling sizes in the tens of thousands.
Aiming to maximize accuracy, this model achieves an average absolute improvement of 3.1\% and 1.5\% over random forests in small-scale and medium-sized datasets, respectively.
Experiments targeting various objectives, including balanced accuracy, cost-sensitive cost, and F1-score, demonstrate the framework's wide applicability and its superiority over contemporary state-of-the-art optimal classification tree methods in small to medium-scale datasets.
}
\end{abstract}


\end{frontmatter}

\section{Introduction}
In recent years, pivotal decision-making tasks across a spectrum of domains, from finance to autonomous driving, have increasingly been automated through advanced machine learning (ML) tools.
Particularly since the advent of deep learning in the last decade, machine learning models have evolved to be larger and structurally more complex.
However, this enhanced predictive performance is not obtained for free.
Nevertheless, models like deep neural networks and ensemble models are inherently black-box, implying they lack transparency in providing the rationale behind their predictions.
Consequently, in many high-stakes decision-making domains, like the judiciary and healthcare, black-box models, despite their impressive predictive capabilities, face challenges in real-world implementation as domain experts tend to resist trusting and adopting models they cannot comprehend \citep{rudin2019}.
The European Union's General Data Protection Regulation mandates the ``right to explanation'' \citep{Parliament2016}, insisting on human-understandable predictive processes in models.

The core of interpretable machine learning is to understand and interpret the decision-making process and results of the model.
Classification tree models such as CART \citep{breiman1984}, C4.5 \citep{quinlan1993} and ID3 \citep{quinlan1986} stand out among various interpretable models for their full transparency and straightforward interpretability.
The paramount advantage of classification trees is their interpretability, making them the preferred choice in high-stakes decision-making areas like healthcare and risk management, where  interpretability  is valued over higher accuracy but less interpretable methods \citep{rudin2019}. 
Consequently, there has been a burgeoning interest in employing classification trees for decision-making problems, attributed to their interpretability, as evidenced in recent scholarly and practical pursuits \citep{bertsimas2020, elmachtoub2020, ciocan2022,aouad2023,zhou2023offline}.
Nonetheless, existing classification trees are encumbered by three significant limitations.

First, there is  a significant gap in prediction accuracy between existing classification tree algorithms and black-box algorithms such as random forest.
The greedy tree-growing based  classification trees like CART and C4.5 frequently yield suboptimal tree structures.
To deal with this problem,  many researchers have focused on developing various models and algorithms to improve the performance of classification tree models, e.g., dynamic programming \citep{linden2023,demirovic2023,Demirovic2022,Lin2020}, mixed-integer programming (MIP)  \citep{Bertsimas2017, verwer2019, aghaei2021, gunluk2021, firat2020,  Subramanian2023},
Boolean satisfiability \citep{Shati2023,Nina2018} and branch-and-bound \citep{mazumder2022quant}.
Although existing optimal classification trees have been demonstrated to surpass heuristic-based ones in accuracy, they still considerably lag behind black-box models.

The second  limitation is the univariate tree structure.
A univariate tree  can be represented as a set of rules, each rule corresponds to a leaf node, and all rules originate from the same root node.
This tree structure, predicated on univariate splits, engenders two primary issues.
First, this structure imposes stringent constraints on learnable rules, leading to overfitting and an inability to accommodate high-dimensional features and significantly heterogeneous datasets.
Second, this tree structure can cause  the replicated subtree problem \citep{bagallo1990,{ragodos2022}}.
Owing to the fragmentation of the sample space by non-overlapping rules, it becomes imperative to learn identical subtrees at various positions within the decision tree like the tree in Figure \ref{fig:14sss}. 
For identical classification outcomes, an increased number of features might be present (refer to EXAMPLE \ref{ex:1}).
Consequently, this structure escalates the computational complexity of the optimal classification tree formulated by MIP methods because the univariate tree requires learning deeper trees to fit the dataset. 

Third, while MIP-based optimal classification trees \citep{bertsimas2007,Bertsimas2017,aghaei2021, gunluk2021, firat2020, verwer2019} typically endeavor to minimize classification errors, employing metrics such as F1 score and balanced accuracy is more apt for imbalanced datasets.

To address the aforementioned limitations of classification trees, this paper introduces a novel MIP formulation, grounded in multivariate Boolean rules, to learn the optimal classification tree, termed BooleanOCT.
Unlike univeriate classification trees,  BooleanOCT facilitates node splitting through the application of multivariate Boolean rules.
The flexibility of these split rules enables the generation of hypercubes in multiple dimensions, encompassing both closed and open forms.
Consequently, BooleanOCT effectively overcomes the challenge of replicated subtrees and adeptly captures the heterogeneity and nonlinear patterns inherent in the data, in the meantime, it maintains a high level of interpretability.

In summary, our research is motivated by the desire to develop both interpretable and accurate classifiers.
The main contributions of this paper are as  follows:
\begin{enumerate}
  \item 
      We introduce a novel MIP formulation, leveraging multivariate Boolean rules, to learn the optimal classification tree. This  approach is implemented in an open-source Python package, BooleanOCT.

  \item 
      We have conducted a thorough benchmarking of BooleanOCT against the state-of-the-art models (CART, OCT, FlowOCT, StreeD) on a sample of 36 datasets from the UCI machine learning repository.
      The results reveal that BooleanOCT substantially outperforms other leading models in terms of accuracy on small and medium datasets.
      In comparison to CART, BooleanOCT achieves an average absolute improvement of 7.6\% and 3.5\% in small and medium-sized datasets, respectively.

   \item We also provide a comparison of  BooleanOCT  to random forests.  BooleanOCT has improved the average accuracy by 3.1\% and 1.5\% on the  small-scale and medium-scale datasets, respectively.
       To the best of our knowledge, BooleanOCT stands as the first interpretable classification tree model surpassing the performance of the black-box model, random forests.

   \item   Additionally, we demonstrate that by modifying certain constraints, it is feasible to integrate other linear metrics
    (balanced accuracy, cost-sensitive cost)  and nonlinear metrices (F1-score).
    Experimental evidence across these three application domains underscores the flexibility of BooleanOCT, while performing on par with or better than the state of the art.
       
The structure of the paper is as follows. 
Section \ref{sec:related work} provides a review of the existing literature on interpretable machine learning  models and optimal classification trees.
In Section \ref{sec:BooleanOCT}, the BooleanOCT model is introduced.
Section \ref{sec:imbalanced} focuses on the development of several variants of the BooleanOCT model. These variants are specifically designed to handle imbalanced datasets, which is a common challenge in binary classification problems.
In Section \ref{sec:experiments}, computational experiments are presented to demonstrate the effectiveness of the proposed methods. 
Finally, Section \ref{sec:conclusion} concludes the paper with pointers for future work.

\end{enumerate}

\section{Related Work}
\label{sec:related work}
We  begin this section by examining recent studies on interpretable machine learning models, followed by a discussion on the latest advancements in optimal classification tree models.

\subsection{Interpretable Machine Learning}
To enhance human comprehension, different types of interpretable models have been developed. 
There are two main approaches to achieving interpretability. 
The first approach involves developing a highly accurate black-box model and then employing post hoc explainers to generate interpretations for it.
Prominent among these explainers are  SHapley Additive exPlanation (SHAP) \citep{Lundberg2017} and Local Interpretable Model-agnostic Explanations (LIME) \citep{Ribeiro2016}.
These tools strive to open the black-box models by estimating the contribution of individual features to specific predictions.
Nevertheless, recent years have witnessed significant concerns regarding post hoc methods.
Issues such as ambiguity, inconsistency, and multiple explanations have been raised by researchers  \citep{Chen2022,rudin2019,laugel201911}.
These explanations can sometimes be manipulated easily  \citep{Slack2020} or show significant variations when parameter settings are changed \citep{ross2017}. These drawbacks highlight the need for more reliable and robust interpretability techniques.

Consequently, to attain interpretability with substantial practical relevance, 
the second approach, which involves building a model that is inherently interpretable and understandable on its own, has gained increasing focus.
This category encompasses models like sparse linear models \citep{ustun2016}, generalized additive models \citep{Lou2012}, rule-based models \citep{ragodos2022,Balvert2024}, decision trees \citep{breiman1984}, among others.
A key advantage of this approach is the absence of any ambiguity in the explanations, as they are intrinsically intertwined with the decision-making processes. 
While recent cutting-edge research in interpretable machine learning has enhanced the capabilities of these models, allowing them to sometimes match the performance of black-box models, it is more commonly observed that interpretable models fall behind black-box models when it comes to complex tasks. This is primarily because interpretable models have to adhere to the specific interpretability requirements of the domain and the user \citep{Wang2021}.

\subsection{Optimal Classification Trees}

The Classification and Regression Trees (CART) method, initially proposed by \cite{breiman1984}, is widely recognized as the leading approach for classification using decision tree methods. CART uses a top-down strategy to create partitions.
However, a notable limitation of this approach, as well as other popular methods like C4.5 (Quinlan 1993) and ID3 (Quinlan 1986), is their inherent greediness. In these methods, each split in the decision tree is determined independently, without considering the potential consequences of subsequent splits. Consequently, the resulting trees may fail to effectively capture the dataset's underlying characteristics, leading to suboptimal performance in classifying future instances.
Another limitation associated with heuristic-based approaches is their inherent challenges when it comes to training multivariate or oblique decision trees. In such cases, node splits  require consideration of multiple variables or hyperplanes. 
Furthermore, due to their greedy nature, these heuristic methods face difficulties in incorporating essential global properties, such as model sparsity and cost-sensitivity.

In response to these challenges, many scholars are dedicated to developing optimal classification trees that aim to globally optimize a specific objective within a predefined maximum size \citep{Bertsimas2017}. 
In this section, we specifically focus on recent developments in MIP-based optimal classification trees and dynamic programming-based optimal classification trees. For a comprehensive overview of the advancements in optimal classification tree algorithms, interested readers can refer to the latest literature review \citep{carrizosa2021,{costa2023recent}}.

\subsubsection{MIP-based Optimal Classification Trees}
The utilization of top-down induction and pruning in the CART algorithm did not stem from the belief that this approach was intrinsically superior. Rather, it was motivated by practical considerations at the time, primarily because finding an optimal tree presented an NP-hard problem  \citep{laurent1976}.
However, with the ongoing advancements in software and hardware, specific MIP solvers, such as Gurobi  and CPLEX, now possess the capability to efficiently tackle MIP problems on a large scale. Consequently, the prospect of learning optimal classification trees with restricted depth has become feasible.

\cite{Bertsimas2017}  developed two mixed-integer programming-based models for optimal sparse classification trees: OCT (for axes-aligned splits) and OCT-H (for multivariate splits). 
These models showed average absolute improvements in out-of-sample accuracy over CART of 1-2\% and 3-5\%, respectively. Subsequently, they applied robust optimization techniques to enhance OCT's robustness \citep{bertsimas2019} and stability \citep{bertsimas2022stable}.
However, OCT and OCT-H face intractability issues due to the exponential growth in the number of binary variables and constraints, approximately ${\cal O}\left( {{2^D}|\mathcal{I}|} \right)$, where $D$ is the tree's depth, and $|\mathcal{I}|$ is the number of training dataset samples. Consequently, efforts have been made to improve the efficiency of these approaches.

\cite{verwer2019} introduced BinOCT, a binary linear program formulation with significantly fewer decision variables and constraints than OCT. 
\cite{aghaei2021} recently proposed FlowOCT, a strong max-flow-based MIP formulation for optimal binary classification trees, demonstrating its superiority in linear optimization relaxation over BinOCT and OCT. 
\cite{firat2020} employed column generation techniques, capable of handling large datasets, while still producing competitive results compared to CART. \cite{Subramanian2023} presented an independent path-based MIP formulation, outperforming existing models (OCT, FlowOCT, BinOCT, and CART) on most small to medium-scale datasets.

While these models, along with other state-of-the-art MIP-based models \citep{gunluk2021,carrizosa2021,zhu2020,hua2022scalable}, have made significant contributions in terms of scalability and/or prediction accuracy, they still have some limitations. Firstly, despite MIP-based optimal classification trees being proven more accurate than heuristic-based classification trees on small to medium-scale datasets, they still fall behind black-box models like Random Forests. For instance, on average, Random Forests exhibit approximately 6\%, 5\%, and 3\% higher accuracy than CART, OCT, and OCT-H at a depth of 4, respectively \citep{Bertsimas2017}.
Another limitation is that most MIP-based models often require several hours to learn trees with a depth less than 4, particularly when applied to medium to large-scale datasets. Similarly, alternative methods such as constraint programming \citep{verhaeghe2020learning} and Boolean satisfiability \citep{Shati2023,Nina2018,avellaneda2020efficient}  also face scalability challenges, as they rely on generic solvers, as observed in the case of MIP-based models \citep{costa2023recent}.

\subsubsection{Dynamic Programming-based Optimal Classification Trees}
Rather than relying on the generic solvers, another group of works focuses on utilizing dynamic programming approaches to learn optimal classification trees, achieving orders of magnitude better scalability \citep{aglin2020, Demirovic2022, Lin2020, linden2023}.

\cite{nijssen2007mining} introduced the pioneering dynamic programming approach for optimal classification trees, known as DL8. This method formulates the task of classification tree induction as a problem akin to frequent itemset mining.
Subsequently, they extended DL8 to accommodate a broader set of constraints, including privacy preservation \citep{nijssen2010optimal}. 
Building upon these ideas, \cite{aglin2020} proposed DL8.5, an algorithm that enhances scalability through sophisticated caching techniques and branch-and-bound search methods, effectively reducing the search space. 
MurTree \citep{Demirovic2022}  improves scalability through the integration of a special solver designed for trees with a depth of two, the implementation of a similarity lower bound, and the incorporation of various advanced techniques.
Blossom \citep{demirovic2023}, an anytime algorithm, which further enhances the scalability compared to MurTree.
StreeD, developed by \citep{linden2023}, represents the most recent advancement in this research domain, generalizing previous dynamic programming approaches into a versatile framework capable of optimizing various combinations of separable objectives and constraints.

In parallel, a separate line of specialized algorithms emerged, starting with OSDT \citep{hu2019}. 
Building on previous work on the CORELS algorithm \citep{angelino2018learning}, OSDT employs an explicit branch-and-bound approach to efficiently explore and prune the search space.
An extension known as GOSDT \citep{Lin2020} has been developed to expand OSDT's capabilities, allowing it to consider objective measures beyond accuracy and improving its ability to handle continuous attributes.
However, these algorithms often demand impractical amounts of computational time and memory when attempting to find optimal or near-optimal trees for large-scale datasets. 
\cite{mctavish2022fast}  recently introduced a compelling solution through intelligent guessing strategies derived from  black-box models. 
These strategies are applicable to any branch-and-bound-based decision tree algorithm,  significantly reducing runtime by several orders of magnitude.

Based on experimental results from previous studies \citep{hu2019,aglin2020,Lin2020, Demirovic2022, demirovic2023,mctavish2022fast}, it can be concluded that most dynamic programming-based optimal classification trees offer superior scalability and accuracy compared to MIP-based univariate optimal classification trees. 
Notably, Blossom, StreeD, and MurTree can find optimal trees within seconds or minutes when the depth is less than 5 on large-scale datasets. 
Consequently, it can be inferred that Blossom, StreeD, and MurTree outperform existing optimal classification tree models such as DL8.5, BinOCT, GOSDT, OSDT, and bsnsing.
 
However, while some results suggest that dynamic programming-based optimal classification trees \citep{mctavish2022fast,Demirovic2022} can occasionally compete with black-box models like Random Forests, there is still a significant performance gap in most cases when compared to Random Forests. 
Moreover, other tree structures like multivariate classification trees \citep{dunn2018optimal, {zhu2020},{boutilier2023optimal}} and multiway-split decision trees \citep{Subramanian2023} have shown significant accuracy improvements compared to univariate classification trees. However, existing dynamic programming-based models are primarily designed for univariate classification trees.

\subsection{Novelty of BooleanOCT}
To the best of my knowledge, two models are highly relevant to our paper: OCT-H \citep{Bertsimas2017} and bsnsing \citep{Liu2022}. 
The comparison between BooleanOCT and these two models, OCT-H and bsnsing, is as follows:

OCT-H represents the first MIP-based model designed for learning optimal classification trees with hyperplanes.
Remarkably, OCT-H achieves comparable performance to random forests when the number of classes is less than 3 and the number of features is less than 25 \citep{Bertsimas2017}. 
However, OCT-H and other state-of-the-art multivariate classification trees  \citep{blanco2023,{zhu2020},boutilier2023optimal} lack interpretability for the hyperplane splits compared to univariate classification trees. 
In contrast, BooleanOCT generates node splits using multivariate Boolean rules, ensuring the same level of interpretability as univariate classification trees.

The bsnsing model, proposed by \cite{Liu2022}, stands out from other optimal classification tree models by not aiming to optimize the entire tree while maintaining the recursive partitioning scheme. 
Instead, it employs heuristic algorithms to generate splits using multivariate Boolean rules. 
Essentially, bsnsing can be considered a classification tree model based on heuristic algorithms. 
The primary advantage of bsnsing compared to other optimal classification trees (DL8.5, OSDT, GOSDT) is its rapid training speed,  while maintaining competitive prediction accuracy. 
Two main differences between BooleanOCT and bsnsing are worth noting. 
First, the split rules based on multivariate Boolean rules in BooleanOCT are more flexible and versatile than those in bsnsing, as elaborated in Remark \ref{remarkbsning}. 
Second, BooleanOCT utilizes MIP to optimize the whole tree under a global objective, surpassing heuristic methods by gaining a superior balance between predictive performance and model complexity.

\section{BooleanOCT}
\label{sec:BooleanOCT}

In this section, we lay out the MIP formulation to learn the optimal classification trees with multivariate Boolean split rules, which optimizes the split strategies and trades off the misclassification errors and model complexity.
We begin by providing some preliminaries on the classification tree models and defining the BooleanOCT  in Section \ref{pre2.1}. We then present our MIP formulation in Section \ref{sec:2.2}.

\subsection{Model Description}
\label{pre2.1}
Let $\mathcal{F}$ represent a set of features, and $\mathcal{K}$ be a set of labels used to describe instances. 
We are given a training dataset $S=\left\{\left(\mathbf{x}_{i}, y_{i}\right)\right\}_{i=1}^{|\mathcal{I}|}$ , which contains $|\mathcal{I}|$ instances. 
Each instance $i \in \mathcal{I}$ consists of a label $y_{i} \in \{0,1, \ldots, |\mathcal{K}|-1\}$ and a set of $|\mathcal{F}|$ binary features, represented as the vector $\boldsymbol{x}_{i} \in\{0,1\}^{|\mathcal{F}|}$. 

BooleanOCT is a classification tree designed to predict the label $y_i$ for a given instance $i$ using its features $\mathbf{x}_{i}$ by evaluating a set of splits. 
The split strategy at a branch node takes the form of \textbf{IF} ``instance satisfies formula'' \textbf{THEN} ``branch to its left child node'' \textbf{ELSE} "branch to its right child node''. Here, the formula is a Boolean function that operates on a subset of the features.
In the context of BooleanOCT, a split at branch node $t$ takes the form of:

\begin{equation}
\label{query1}
\text{Is} \quad \sum\limits_{f \in \mathcal{S}_t} x_{if} \leq b_t?
\end{equation}
where $\mathcal{S}_t$ represents the set of features selected in branch node $t$ to generate the split, and $b_t$ ($0 \leq b_t \leq |\mathcal{S}t|$) is an integer. For instance $i$, if the condition $\sum\limits_{f \in \mathcal{S}_t} x_{if} \leq b_t$ holds at branch node $t$, then this instance will proceed to the right child node of node $t$. Otherwise,  this instance will branch to branch node $t$'s left child node.

\begin{figure}[h]
		\centering
		\includegraphics[scale=0.5]{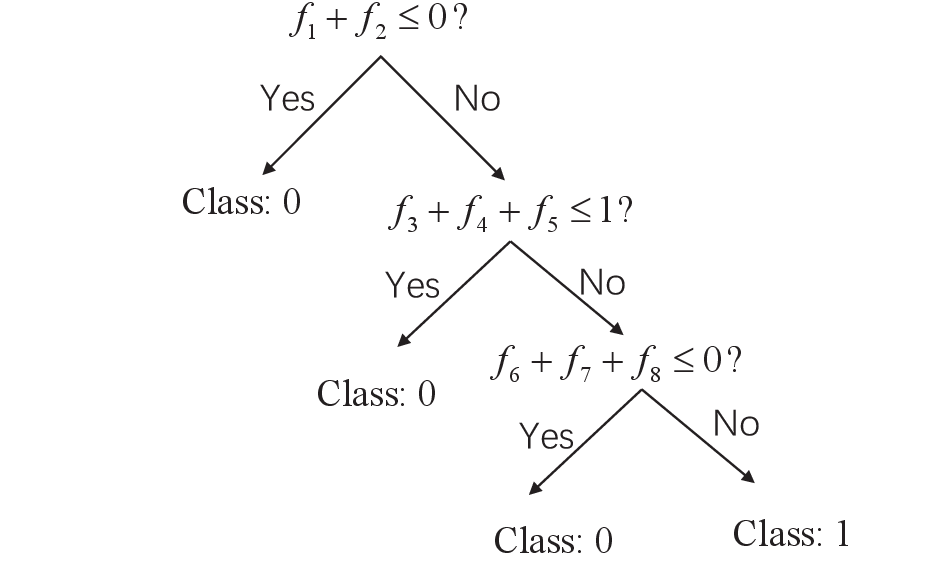}
		\caption{The BooleanOCT  tree structure of the application   conditions for professor's position.}
		\label{fig:condtionpt}
\end{figure}

In certain classification problems, the classification rules align with the multivariate Boolean split strategy employed in BooleanOCT. 
For instance, consider the application requirements for professorial titles in a Chinese university. 
These requirements are presented in Table \ref{condtionpt} and can be represented as multivariate Boolean formulas.
 To illustrate, an applicant must meet at least one condition in the `project' category, expressed as
 ``$f_1 + f_2 \leq 0$ ?''.
If $f_1 + f_2 = 0$, then 
the applicant does not meet the project criteria. 
Similarly, conditions related to scientific research achievements and social work can also be represented as multivariate Boolean formulas.
Consequently, we can utilize multivariate Boolean rules to represent Table \ref{condtionpt}'s conditions in a BooleanOCT tree structure depicted in Figure \ref{fig:condtionpt}.

\begin{table}[H]
\centering
\caption{Application   conditions for professor's position}   
\newcommand{\tabincell}[2]{\begin{tabular}{@{}#1@{}}#2\end{tabular}}
\setlength{\tabcolsep}{1mm}{
\begin{tabular}{lll}
\hline                                                                                                                                                                                                                                                                                                                                                                                                                                                
\multirow{2}{*}{Project}                          & \multirow{2}{*}{\tabincell{l}{Meet at least\\ one of   them}} & 
\tabincell{l}{1. Hosted and completed a national\\ level scientific research project.}                                                                                                                                                                                                                                                                                                                                                     \\
                                                  &                                              & 
\tabincell{l}{2. Led   horizontal scientific research projects\\ with a cumulative funding of 500000   yuan.}                                                                                                                                                                                                                                                                                                                                 \\\hline   
\multirow{3}{*}{\tabincell{l}{Scientific research\\ achievements}} & \multirow{3}{*}{\tabincell{l}{Meet at least \\two of   them}} & 
3. Published 3 papers in A-level journals.                                                                                                                                                                                                                                                                                                                                                                                \\
                                                  &                                              & 
4. Editor in chief of a national level   textbook.                                                                                                                                                                                                                                                                                                                                                                            \\
                                                  &                                              & 
\tabincell{l}{5.   Received one first prize or higher (ranked top three) in \\provincial and   ministerial level humanities and social sciences\\ (teaching achievements), one   second prize (ranked top two) \\in provincial and ministerial level humanities   and social\\ sciences (teaching achievements), or one third prize (ranked   first)\\ in provincial and ministerial level humanities \\and social sciences   (teaching achievements).} \\\hline   
\multirow{3}{*}{Social work}                      & \multirow{3}{*}{\tabincell{l}{Meet at least\\ one of   them}} & 
\tabincell{l}{6. Serve as department head,   homeroom teacher, counselor, \\credit system supervisor, or innovation class   supervisor \\for more than 1 year.}                                                                                                                                                                                                                                                                                \\
                                                  &                                              & 
\tabincell{l}{7.   Engaged in teaching, scientific research, \\equipment management work for more   than 1 year.}                                                                                                                                                                                                                                                                                                                            \\
                                                  &                                              & 
\tabincell{l}{8.   Undertook social work such as supporting education,\\ going to rural areas,   assisting Tibet, assisting foreign countries,\\ poverty alleviation, and post   disaster reconstruction as required by \\ schools or other higher-level   departments.}      
\\\hline                                                                                                                                                                     
\end{tabular}}
\label{condtionpt}
\end{table}

The BooleanOCT tree structure not only learns the rules like the application conditions for professor’s position problems with  multivariate Boolean rules, but also can significantly reduce the complexity of the tree structure and improve the fitness.
The following example is to illustrate the BooleanOCT can better fit some datasets while reducing the complexity.

\begin{example}
\label{ex:1}
Consider a binary classification problem with 10 samples and 5 features. The specific details are provided in  Table \ref{table:exampele1}.
\end{example}
\begin{table}[H]
\centering
  \caption{A binary classification problem with 10 samples and 5 features, \\the last column indicating whether the sample is positive or negative.}
\begin{tabular}{lllllllllll}
\hline
instance & $e_1$ & $e_2$ & $e_3$ & $e_4$ &$e_5$ & $e_6$ &$e_7$ & $e_8$ & $e_9$ & $e_{10}$ \\\hline
$f_1$    & 0  & 0  & 0  & 0  & 1  & 1  & 1  & 1  & 1  & 1   \\
$f_2$    & 0  & 0  & 1  & 1  & 0  & 1  & 0  & 1  & 1  & 1   \\
$f_3$    & 0  & 1  & 0  & 1  & 0  & 0  & 1  & 1  & 1  & 1   \\
$f_4$    & 1  & 0  & 0  & 0  & 1  & 1  & 0  & 0  & 0  & 1   \\
$f_5$    & 0  & 1  & 0  & 1  & 0  & 1  & 0  & 1  & 0  & 1   \\\hline
Class    & 0  & 0  & 0  & 1  & 0  & 1  & 1  & 1  & 1  & 1   \\\hline
\end{tabular}
\label{table:exampele1}
\end{table}

A valid and optimal decision set for this data is
\begin{equation*}
\begin{array}{lll}
\text{\textsf{if}} & f_1 \wedge f_2 & \text{\text{\textsf{then}} the predicted label is 1,} \\
\text{\textsf{else if}} & f_1 \wedge f_3 & \text{\text{\textsf{then}} the predicted label is 1,} \\
\text{\textsf{else if}} & f_2 \wedge f_3 & \text{\text{\textsf{then}} the predicted label is 1,} \\
\text{\textsf{else}} &  & \text{\text{\textsf{then}} the predicted label is 0.} \\
\end{array}
\end{equation*}

The decision rules within the decision set are such that if at least two out of the three features ($f_1$, $f_2$, and $f_3$) are true, the instance will be predicted as label 1. Consequently, this binary classification problem can be expressed as a classification tree based on multivariate Boolean rules with a depth of 1. The split at the root node takes the form of:

\begin{equation}
\label{queryroot}
\text{Is} \quad f_1 + f_2 +f_3 \leq 1?
\end{equation}

\begin{figure}[H]
\centering
    \begin{minipage}[t]{0.4\linewidth}
        \centering
        \includegraphics[width=\textwidth]{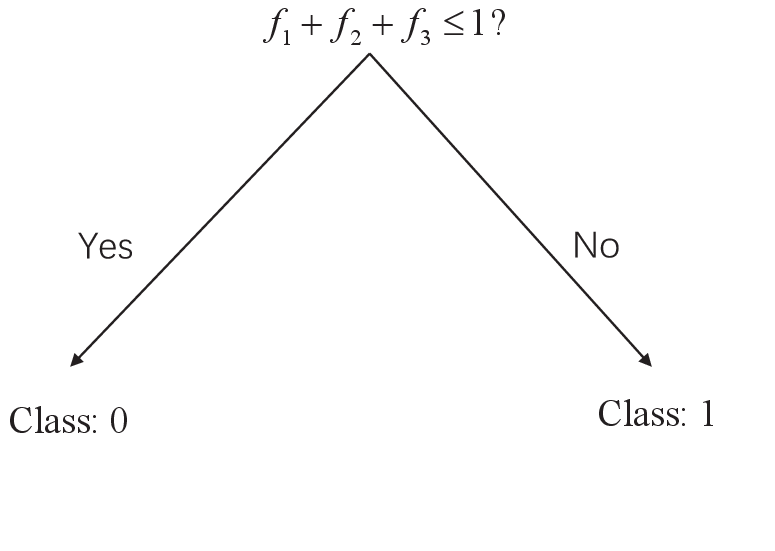}
        \centerline{(a) The BooleanOCT tree structure.}
    \end{minipage}%
    \begin{minipage}[t]{0.6\linewidth}
        \centering
        \includegraphics[width=\textwidth]{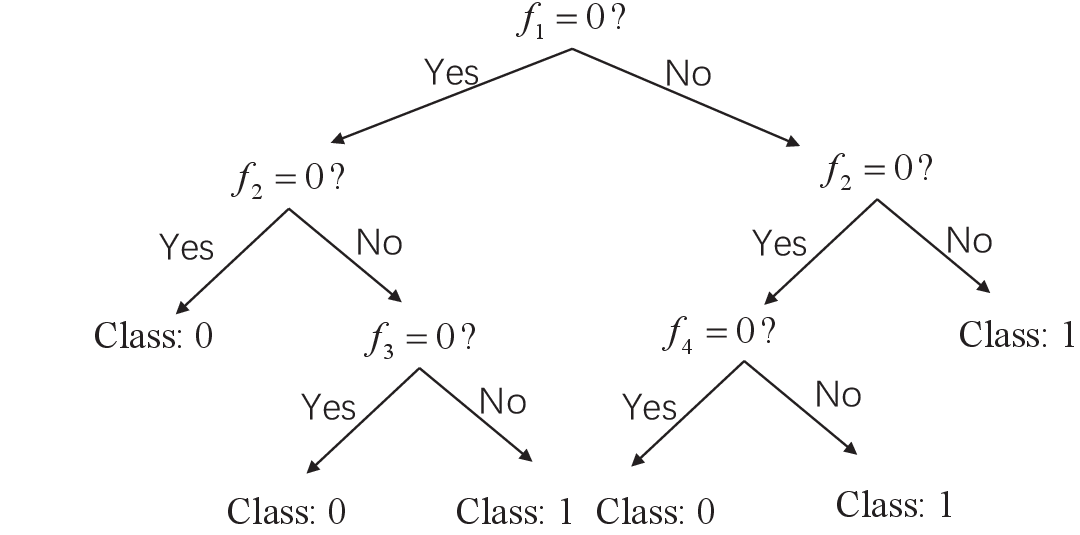}
        \centerline{(b) The CART tree structure.}
    \end{minipage}
    \caption{The tree structures of EXAMPLE \ref{ex:1}.}
\label{fig:example1b}
\end{figure}

In the case of instance $i$, if $f_1 + f_2 +f_3 \leq 1$, then instance $i$ will be directed to leaf node 2 and assigned a label of 0. Conversely, if 
$f_1 + f_2 +f_3 > 1$, instance $i$ will be directed to leaf node 3 and assigned a label of 1. The tree structure for this binary classification problem, based on multivariate Boolean rules, is depicted in Figure \ref{fig:example1b} (a).

On the other hand, if we employ univeriate classification models such as CART to fit this dataset, CART will produce a tree with a depth of 3, as illustrated in Figure \ref{fig:example1b} (b). 
This tree is considerably more complex than  BooleanOCT  and could lead to overfitting due to its 5 branch nodes and the selection of 4 features for generating the splits.

MIP-based models like OCT and BinOCT are NP-hard problems, and the number of binary variables in these models scales with ${\cal O}\left( {2^D|\mathcal{I}|} \right)$, where $D$ represents the depth of the tree, and $|\mathcal{I}|$ denotes the number of samples in the training dataset. 
Consequently, when we compare BooleanOCT with MIP-based optimal classification trees (OCT and BinOCT) for datasets like EXAMPLE \ref{ex:1}, the latter will necessitate approximately four times more binary decision variables than BooleanOCT. 
Consequently, BooleanOCT can generate trees with shallower depths, which accelerates the solving speed and mitigates the risk of overfitting.

\begin{remark}
Let $N_{max}^d$ be the maximal number of features among all the branch nodes at depth $d$ in BooleanOCT.
 If, for all $d$ in $\mathcal{D}$, $N_{max}^d = 1$, then BooleanOCT is a univariate classification tree with a depth of $D$.
In BooleanOCT, the split strategy at a branch node is a  Boolean function applied to a subset of features. 
This node can be equivalently represented as a subtree, with its depth being equal to the number of features selected at that node. 
Consequently, we can replace the Boolean rules at each branch node in BooleanOCT with a subtree. 
Thus, BooleanOCT can be represented as a univariate classification tree with a depth of $D_{binary}$, which is calculated as follows:
\begin{equation}
D_{binary}=\sum\limits_{d=1}^{D} 2^{N_{max}^d-1}.
\end{equation}
\end{remark}

\begin{figure}[H]
		\centering
		\includegraphics[scale=0.5]{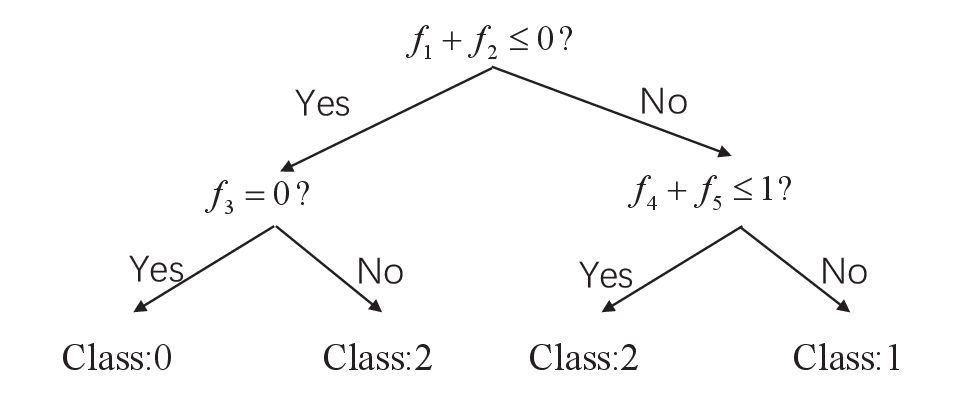}
		\caption{The BooleanOCT  tree structure in EXAMPLE \ref{ex:2}.}
		\label{fig:examp2}
\end{figure}

\begin{figure}[H]
\centering
    \begin{minipage}[t]{0.4\linewidth}
        \centering
        \includegraphics[width=\textwidth]{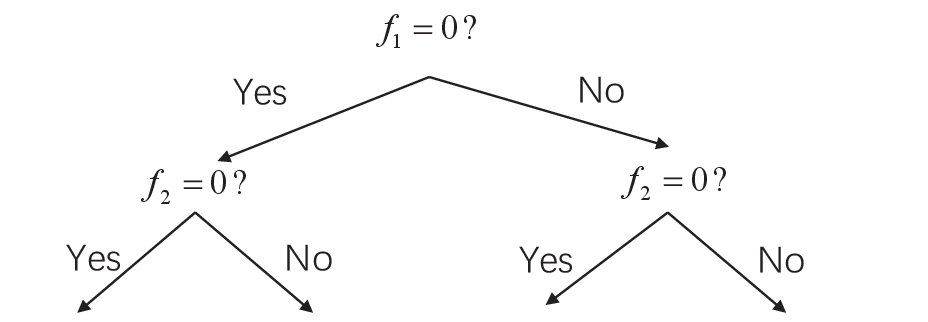}
        \centerline{(a) The branch node ``$f_1 +f_2 \leq 0?$".}
    \end{minipage}%
    \medskip
    \begin{minipage}[t]{0.4\linewidth}
        \centering
        \includegraphics[width=\textwidth]{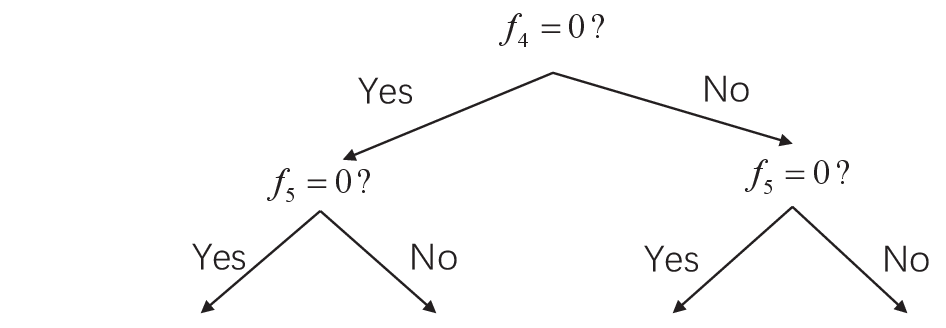}
        \centerline{(b) The branch node ``$f_4 +f_5 \leq 1?$".}
    \end{minipage}
    \caption{The subtrees of the branch nodes with multivariate Boolean rules.}
\label{fig:example2}
\end{figure}

\begin{figure}[H]
		\centering
		\includegraphics[scale=0.5]{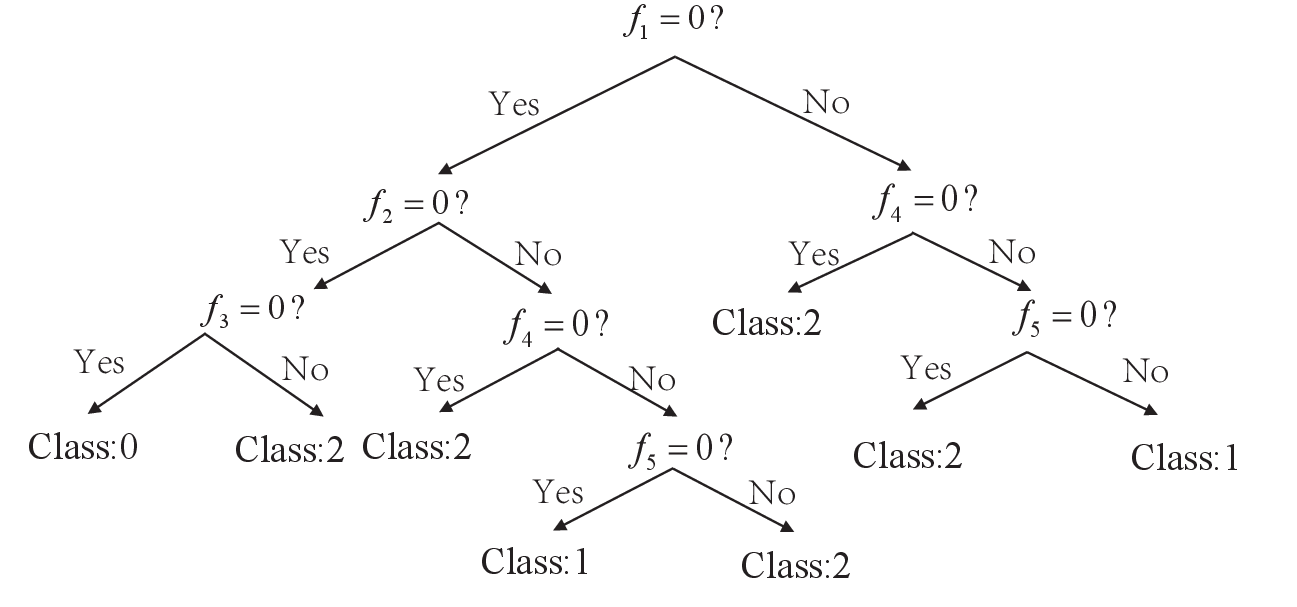}
		\caption{A univariate classification tree for the  BooleanOCT   in EXAMPLE \ref{ex:2}.}
		\label{fig:14sss}
\end{figure}

Let's consider Example \ref{ex:2} as an illustration. 
In the BooleanOCT tree structure, both the root node and the right child node of the root node utilize two features to generate splits. 
Therefore, we can represent these two branch nodes as two univariate classification subtrees, each with a depth of 2, as shown in Figure \ref{fig:example2}. These two univariate classification trees can replace the multivariate Boolean rules in Figure \ref{fig:examp2}, resulting in a  univariate classification tree with a depth of 4, as depicted in Figure \ref{fig:14sss}.
It's worth noting that in the univariate classification tree (Figure \ref{fig:14sss}), the positions of $f_1$ and $f_2$ can be interchanged, and similarly, the positions of $f_4$ and $f_5$ can also be exchanged. 
Consequently, the BooleanOCT in Figure \ref{fig:examp2} can be equally represented as eight different binary classification trees, all with the same classification rules.

\begin{example}
\label{ex:2}
Consider a BooleanOCT with a depth of 2. Its tree structure is shown in Figure \ref{fig:examp2}.
\end{example}

\subsection{Problem Formulation}
\label{sec:2.2}
In this section, we will design a MIP model for learning an optimal multi-class classification tree with a predetermined depth denoted as  $D$.
While much of the formulation is based on the OCT model presented by \cite{Bertsimas2017}, our approach does not impose the restriction of using at most one feature at each branch node.

Let $\mathcal{F}$ represent the set of features and $\mathcal{K}$ denote the set of labels used to describe an instance. We have a training dataset, denoted as $S=\left\{\left(\mathbf{x}_{i}, y_{i}\right)\right\}_{i=1}^{|\mathcal{I}|}$, which contains $|\mathcal{I}|$ instances. 
Each row  $i \in {I}$  consists of  a label  $y_{i} \in\{0,1, \ldots, |\mathcal{K}|-1\}$ and 
$|F|$  binary features  which we collect in the vector  $\boldsymbol{x}_{i} \in\{0,1\}^{|F|}$.
To formulate the MIP model effectively, we need to establish the necessary notations and decision variables, which are summarized in Table \ref{tab:para} and Table \ref{tab:variable}, respectively.

\begin{table}[H]
  \centering
  \caption{Summary of notation}
    \begin{tabular}{ll}
        \hline
Notation & Definition\\
\hline
   Set & \\
$\mathcal{I}$ & \text { Set of training data } \\
$\mathcal{F}$ & \text { Set of features } \\
$\mathcal{K}$ & \text { Set of labels } \\
$\mathcal{B}$ & \text { Set of branch nodes in the decision tree } \\
$\mathcal{L}$ & \text { Set of leaf nodes in the decision tree } \\
$\mathcal{A}(t)$ & \text { Set of ancestors of node } $t$ \text { in the decision tree } \\
$\mathcal{A}_R(t)$ & \text { Set of right-branch ancestors of node } $t$ \\
$\mathcal{A}_L(t)$ & \text { Set of left-branch ancestors of node } $t$ \\
$\mathcal{L}_P(t)$ & \text { Set of potential parent nodes of leaf node } $t$ \\
$\mathcal{S}_t$   & \text  {Set of features selected in branch node $t$ to generate the split}\\

Index & \\
$i$ & \text { Instance in training data, } $i \in \mathcal{I}$ \\
$f$ & \text { Feature in training data, } $f \in \mathcal{F}$ \\
$k$ & \text { Label in training data, } $k \in \mathcal{K}$ \\
$t $& \text { Node of tree, } $t \in \mathcal{B} \cup \mathcal{L}$ \\
$p(t)$ & \text { Parent of node } $t$ \\
Parameter & \\
$D$ & the depth of the tree\\
$\alpha$  &the complexity parameter\\
$S_{\text {min }}$ &minimum number of instances at each leaf node\\
$F_{max}$ & the maximum number of features used in a branch node\\
\hline
\end{tabular}%
\label{tab:para}%
\end{table}%

\begin{table}[H]
  \centering
  \caption{Summary of decision variables}
    \begin{tabular}{ll}
        \hline
Notation & Definition\\
\hline
 $a_{t f} $ & if feature   $f $  is selected to split at branch node   $t$ $(=1) $  or not   $( =0) $    \\
 $b_{t} $ & the threshold at branching node   $t $\\
  $c_{t k} $ &  if label   $k $  is assigned to leaf node   $t$ ( =1)   or not   $( = 0) $ \\
  $d_{t} $ &  if a split rule is applied to branch node    $t$ ( =1)   or not   $( = 0) $ \\
  $z_{i t} $ &  if instance  $i $  is assigned to leaf node    $t$ ( =1)   or not   $(= 0) $ \\
  $e_{t} $ &  the number of mis-classified instances at leaf node   $t $\\
$l_{t} $ &  if leaf nodes   $t$  has at least   $S_{\min }$  instances    $t$ ( =1)  or not   $(=0)$ \\
 
\hline
\end{tabular}%
\label{tab:variable}%
\end{table}%

\begin{figure}[h]
		\centering
		\includegraphics[scale=0.6]{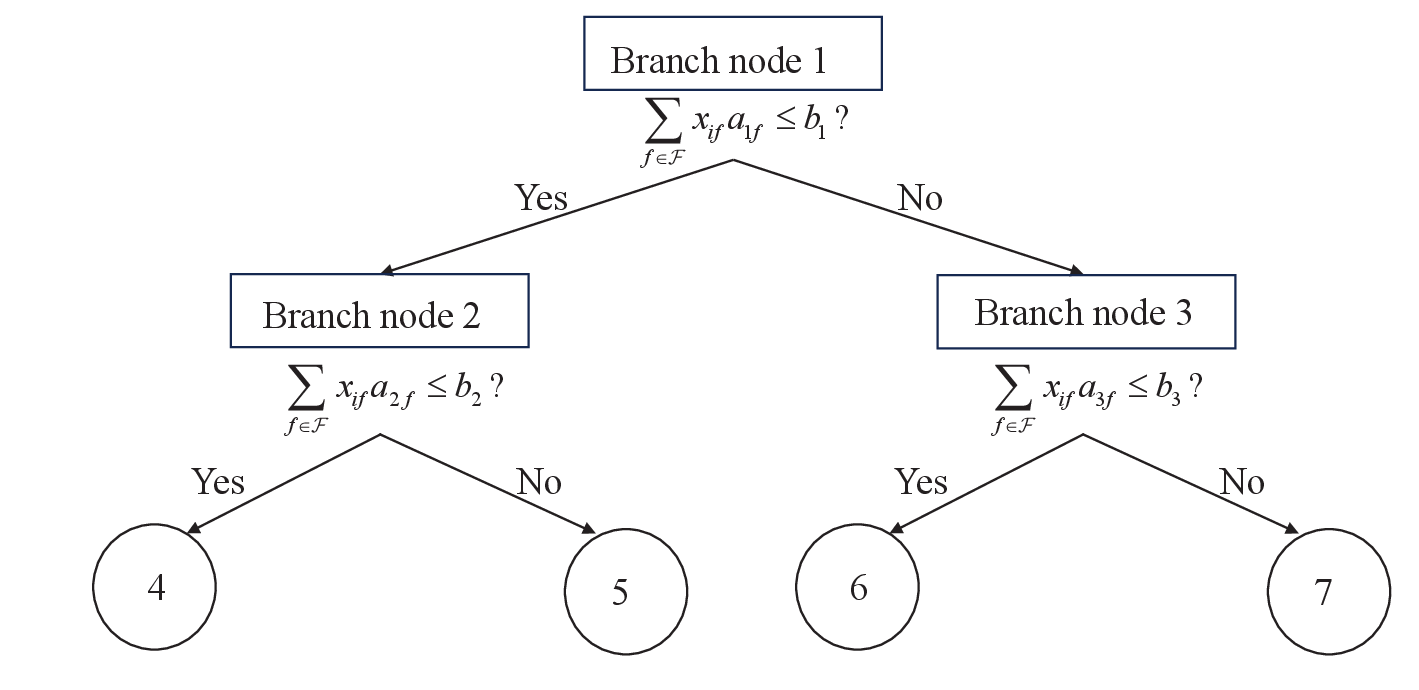}
		\caption{An example of a decision tree with depth $D$=2.}
		\label{fig:1}
\end{figure}

In order to create an MIP formulation that accurately represents the tree structure and enables the search for the optimal solution, it is essential to predefine the maximum depth of the tree, denoted as $D$. This allows us to construct a maximal tree with a depth of $D$, comprising a total of $T$ nodes where $T=2^{D+1}-1$. These nodes are indexed as $t=1, 2, \ldots, T$.
 A decision tree is a maximal tree if every branch node applies a split and every leaf node is located at the maximal depth.  The maximal tree with a depth of 2 is shown in Figure \ref{fig:1}.

The tree consists of two sets of nodes: branch nodes ($t \in \mathcal{B} = {1, \ldots, \lfloor T / 2\rfloor}$) and leaf nodes ($t \in \mathcal{L} = {\lfloor T / 2\rfloor + 1, \ldots, T}$).
Each branch node $t$ follows a split rule in the form of $\sum\limits_{f \in F} x_{if}a_{tf} \leq b_{t}$.
If this inequality is satisfied, instance $i$ proceeds to the left branch; otherwise, it proceeds to the right branch.
When a instance reaches a leaf node, that leaf node provides a class prediction for it.

For the node  $t$, we use  $p(t)$ to refer to its immediate parent node, and $\mathcal{A}(t)$ defines the set of all its ancestors. 
Additionally, we define $\mathcal{A}_{L}(t)$ as the set of ancestors of $t$ where the left branch was followed when forming a path from the root node to $t$. Similarly, $\mathcal{A}_{R}(t)$ represents the ancestors whose right branch was followed on the path from the root node to $t$, and it follows that $\mathcal{A}(t) = \mathcal{A}_{L}(t) \cup \mathcal{A}_{R}(t)$.
For example, considering the tree shown in Figure \ref{fig:1}, we have $\mathcal{A}(6)=\{1,3\}, \mathcal{A}_{L}(6)=\{3\}$, and  $\mathcal{A}_{R}(6)=\{1\}$.

To establish the hierarchical structure of the tree, we introduce the binary variable $d_{t}$, which tracks whether a split occurred at branch node $t$. Specifically, $d_{t}=1$ when a split is made, and $d_{t}=0$ otherwise.
It's important to note that a split at branch node $t$ is restricted if its parent node $p(t)$ does not allow a split. However, since the root node has no parent node, we do not impose any constraints on the root node.
These conditions are enforced through the following constraints:

\begin{equation*}
\label{dp}
\begin{array}{ll}
d_{t} \leq d_{p(t)} & \forall t \in \mathcal{B} \backslash\{1\}\\
d_{t} \in\{0,1\} & \forall t \in \mathcal{B} \\
\end{array}
\end{equation*}

Next, let's introduce the constraints that express the relationships between branch nodes and leaf nodes.

\begin{equation*}
\begin{array}{ll}
l_{t} \leq \sum\limits_{s \in \mathcal{L}_p(t)} d_s & \forall t \in \mathcal{L}\\
Dl_{t} \geq \sum\limits_{s \in \mathcal{L}_p(t)} d_s & \forall t \in \mathcal{L}\\
l_{t} \in\{0,1\} & \forall t \in \mathcal{L} \\
\end{array}
\end{equation*}
where $\mathcal{L}_p(t)$  represents the set of potential parent branch nodes for leaf node $t$. For example, in the tree depicted in Figure \ref{fig:1}, we have $\mathcal{L}_p(4) = \{1,2\}$ and $\mathcal{L}_p(5) = \{2\}$.
The first constraint ensures that there is at least one potential parent branch node when a leaf node $t$ exists.
The second constraint states that a leaf node $t$ must exist if its set of potential parent branch nodes is not empty.

At branch node $t \in \mathcal{B}$, we identify a split using variables $a_{tf}$, $b_{t}$, and $d_{t}$. 
In contrast to existing MIP formulations \citep{Bertsimas2017,aghaei2021,verwer2019}, we allow for the possibility of using multiple features to generate a split at branch node $t$.
We introduce a binary variable $a_{tf}$ that indicates whether feature $f$ is selected to be used for splitting at branch node $t$ ($a_{tf} = 1$) or not ($a_{tf} = 0$). Consequently, $\sum\limits_{f \in \mathcal{F}} a_{t f}$ represents the number of features used at branch node $t$ and should not exceed the predetermined value $F_{max}$, which is the maximum number of features allowed in a branch node.
Additionally, we use an integer variable $b_{t}$ to represent the split threshold value at branch node $t$, with its range of values between 0 and $F_{max}-1$.
Furthermore, if branch node $t$ does not apply a split, then no feature is used at this branch node, and the split threshold value should be zero, which can be expressed as: $d_t=0 \Rightarrow \sum\limits_{f \in \mathcal{F}} a_{tf}=0 \vee b_t=0$.
These conditions are enforced through the following constraints:

\begin{equation*}
\begin{array}{ll}
\sum\limits_{f \in \mathcal{F}} a_{t f} \leq F_{max}d_{t} & \forall t \in \mathcal{B} \\
0 \leq b_{t} \leq (F_{max}-1)d_{t}      & \forall t \in \mathcal{B}\\
a_{tf} \in\{0,1\} & \forall t \in \mathcal{B}, \forall f \in \mathcal{F} \\
b_{t} \in\{0,1, ..., F_{max}-1\} & \forall t \in \mathcal{B} \\
\end{array}
\end{equation*}

Now we have formulated the model such that it captures the structure of the tree. 
Next, let's introduce the binary variable $z_{it}$ to keep track of instance $i$ within the tree. 
If instance $i$ is assigned to branch node $t$, then $z_{it}=1$; otherwise, $z_{it}=0$.
Additionally, we use the binary variable $l_{t}$ to indicate whether leaf node $t$ contains at least $N_{\min}$ instances ($l_{t}=1$) or not ($l_{t}=0$).
With these variables in place, we can define the following constraints to ensure that each leaf node contains a minimum number of instances, $N_{\min}$:

\begin{equation*}
\begin{array}{ll}
z_{i t} \leq l_{t} & \forall t \in \mathcal{L}, \forall i \in \mathcal{I} \\
\sum\limits_{i \in \mathcal{I}} z_{i t} \geq N_{\text {min }} l_{t} & \forall t \in \mathcal{L},\\
\end{array}
\end{equation*}

For  instance $i$, it can only be assigned to one leaf:
\begin{equation*}
\begin{array}{ll}
\sum\limits_{t \in \mathcal{L}} z_{i t}=1, & \forall i \in \mathcal{I}
\end{array}
\end{equation*}

The structure of the tree is determined by the following multivariate Boolean  splits:
\begin{equation}
\label{fmaxc}
\begin{array}{ll}
\sum\limits_{f \in \mathcal{F}} x_{if} a_{s f} \geq b_{s}+d_s-F_{max}(1-z_{i t}) & \forall i \in \mathcal{I}, \forall t \in \mathcal{L}, \forall s \in \mathcal{A}_{R}(t)\\
\sum\limits_{f \in \mathcal{F}}x_{if}a_{s f} \leq b_{s}+F_{max}(1-z_{i t}) & \forall i \in \mathcal{I}, \forall t \in \mathcal{L}, \forall s \in \mathcal{A}_{L}(t)\\
\end{array}
\end{equation}
where the first constraint  is to ensure that instance $i$ is assigned to the right child leaf $t$ when $\sum\limits_{f \in \mathcal{F}}x_{if}a_{sf} \geq b_{s} + d_s$ for all $s \in \mathcal{A}_{R}(t)$;
the second constraint  is to ensure that instance $i$ is assigned to the left child leaf $t$ when $\sum\limits_{f \in \mathcal{F}}x_{if}a_{sf} \leq b_{s}$ for all $s \in \mathcal{A}_{L}(t)$.

Since $x_{if} \in \{0,1\}$, $\sum\limits_{f \in \mathcal{ F}} x_{if} a_{s f} $ indicates the number of true features in the set of features selected in branch node $s$.
Therefore, the constraints in (\ref{fmaxc}) can also be interpreted as follows: 
if instance $i$'s number of true features in the $\mathcal{S}_t$ is greater than $b_s$, then instance $i$ should be assigned to the right child leaf $t$; otherwise, instance $i$ should be assigned to the left child leaf $t$.

To track the assigned class label for each node, we use the binary variable  $c_{tk}$. 
If label   $k $  is assigned to leaf node   $t$, then  $c_{tk}$=1; otherwise, $c_{tk}$=0.
At most one label should be assigned for each leaf node:

\begin{equation*}
\sum\limits_{k \in \mathcal{K}} c_{tk}=l_{t}, \quad \forall t \in \mathcal{L}
\end{equation*}

Let $M_{kt} $ be the number of instances with label $k$ at leaf node $t$, $N_t$ be the total number of instances in the leaf $t$:
\begin{equation*}
\begin{array}{lll}
M_{k t}=\sum\limits_{i \in \mathcal{I}:y_{i}=1} z_{i t}, & \forall k \in \mathcal{K}, \quad \forall t \in \mathcal{L}\\
    
N_{t}=\sum\limits_{i \in \mathcal{I}} z_{i t}, & \forall t \in \mathcal{L} \\
\end{array}
\end{equation*}
Furthermore, we need to associate each leaf node $t$ with a label, denoted by $c_{t} \in \mathcal{K}$. The leaf label is determined as the most frequent label among all the points assigned to a given leaf:
\begin{equation*}
c_{t}=\underset{k \in \mathcal{K}}{\arg \max }\left\{ M_{k t} \right\} .
\end{equation*}

Hence, we define the optimal misclassification loss $e_{t}$ for each leaf node $t$ as the number of points whose labels differ from the most common class label:
\begin{equation*}
e_{t}=N_{t}-\max _{k \in \mathcal{K}}\left\{M_{k t}\right\}=\min _{k \in \mathcal{K}}\left\{N_{t}-M_{k t}\right\}
\end{equation*}
which is linearized as:
\begin{equation*}
\begin{array}{ll}
e_{t} \geq N_{t}-M_{k t}-n\left(1-c_{k t}\right), & \forall t \in \mathcal{L}, k \in \mathcal{K}, \\
e_{t} \leq N_{t}-M_{k t}, & \forall t \in \mathcal{L}, k \in \mathcal{K}, \\
e_{t} \leq nc_{k t}, & \forall t \in \mathcal{L}, k \in \mathcal{K}, \\
e_{t} \geq 0, & \forall t \in \mathcal{L},
\end{array}
\end{equation*}

With the above definitions in mind, the total misclassification error  is $\sum\limits_{t \in \mathcal{L}} e_{t}$.
Then the objective is written as:
\begin{equation}
\label{cp3poip111}
\min \quad \frac{1}{n} \sum\limits_{t \in \mathcal{L}}e_t + \alpha \sum\limits_{f \in \mathcal{F}}\sum\limits_{t \in \mathcal{B}} a_{tf} 
\end{equation}
where $\alpha$ is the complexity parameter of the tree, which adjusts the trade-off between the complexity of the tree in terms of the number of features and the in-sample accuracy.

With the aforementioned definitions and constraints, the MIP formulation of the BooleanOCT (Optimal Classification Trees based on
multivariate Boolean Rules) is as follows:

\begin{equation}
\label{cp3poip}
  \begin{array}{lll}
    \begin{array}{lll}
    \min \quad \frac{1}{n} \sum\limits_{t \in \mathcal{L}}e_t +\alpha \sum\limits_{f \in \mathcal{F}}\sum\limits_{t \in \mathcal{B}} a_{tf} \\

    \sum\limits_{f \in \mathcal{F}} a_{t f} \leq F_{max}d_{t} & \forall t \in \mathcal{B} & (\ref{cp3poip}a)\\

    0 \leq b_{t} \leq (F_{max}-1)d_{t}      & \forall t \in \mathcal{B} & (\ref{cp3poip}b)\\

    d_{t} \leq d_{p(t)}          &\forall t \in \mathcal{B} \backslash\{1\} & (\ref{cp3poip}c)\\

    \sum\limits_{k \in \mathcal{K}} c_{t k}=l_{t} & \forall t \in \mathcal{L} & (\ref{cp3poip}d)\\
    
    l_{t} \leq \sum\limits_{s \in \mathcal{L}_p(t)} d_s & \forall t \in \mathcal{L} & (\ref{cp3poip}e)\\

    Dl_{t} \geq \sum\limits_{s \in \mathcal{L}_p(t)} d_s & \forall t \in \mathcal{L} & (\ref{cp3poip}f)\\

    \sum\limits_{t \in \mathcal{L}} z_{i t}=1     &\forall i \in \mathcal{I} & (\ref{cp3poip}g)\\

    z_{i t} \leq l_{t}          & \forall i \in \mathcal{I}, \forall t \in \mathcal{L} & (\ref{cp3poip}h)\\

    \sum\limits_{i \in \mathcal{I}} z_{i t} \geq S_{\min } l_{t} & \forall t \in \mathcal{L} & (\ref{cp3poip}i)\\

    \sum\limits_{f \in \mathcal{F}} x_{if} a_{s f} \geq b_{s}+d_s-F_{max}(1-z_{i t}) & \forall i \in \mathcal{I}, \forall t \in \mathcal{L}, \forall s \in \mathcal{A}_{R}(t) & (\ref{cp3poip}j)\\

    \sum\limits_{f \in \mathcal{F}}x_{if}a_{s f} \leq b_{s}+F_{max}(1-z_{i t}) & \forall i \in \mathcal{I}, \forall t \in \mathcal{L}, \forall s \in \mathcal{A}_{L}(t) & (\ref{cp3poip}k)\\
    
    M_{k t}=\sum\limits_{i \in \mathcal{I}:y_{i}=1} z_{i t} & \forall k \in \mathcal{K}, \quad \forall t \in {\mathcal{L}} & (\ref{cp3poip}l)\\
    
    N_{t}=\sum\limits_{i \in \mathcal{I}} z_{i t} & \forall t \in {\mathcal{L}} & (\ref{cp3poip}m)   \\

    e_{t}  \geq N_{t}-M_{k t}-n\left(1-c_{k t}\right) & \forall k \in \mathcal{K}, \quad \forall t \in {\mathcal{L}} & (\ref{cp3poip}n) \\
    
    e_{t} \leq N_{t}-M_{k t}, & \forall t \in \mathcal{L}, k \in \mathcal{K} & (\ref{cp3poip}o) \\
    
    e_{t}  \geq 0 & \quad \forall t \in \mathcal{L}  & (\ref{cp3poip}p)\\
  \end{array}
  \end{array}
\end{equation}

In formulation (\ref{cp3poip}), we have four predefined hyper-parameters to specify: $D$, $F_{max}$, $S_{\min}$, and $\alpha$. 
$D$ represents the maximum depth of the tree.
$F_{max}$ is the maximum number of features used in a branch node.
$S_{\min}$ defines the minimum number of instances required to be allocated to a leaf node.
$\alpha$ serves as the complexity parameter of the tree, which regulates the balance between the tree's complexity in terms of the number of features and its in-sample accuracy.

\begin{remark}
\label{remarkbsning}
When we impose the constraint $\sum\limits_{f \in \mathcal{F}} a_{tf} \leq 1$, it implies that at most one feature should be selected at branch node $t$. 
In this case, the partition strategy at branch node $t$ is equivalent to univariate classification tree models such as CART and OCT.
If we further restrict $\sum\limits_{f \in \mathcal{F}} a_{tf} \leq F_{max}$ and set $b_t=1$, then this partition strategy aligns with the bsnsing model \citep{Liu2022}.
In this paper, $b_t$ is an integer decision variable that is constrained to be less than $F_{max}$. Consequently, the multivariate Boolean rules split strategy in BooleanOCT is more flexible and versatile compared to the strategy in bsnsing.
\end{remark}


\section{Imbalanced Datasets}
\label{sec:imbalanced}
The existing literature on optimal classification trees explores linear metrics such as cost-sensitive cost \citep{linden2023} and balanced accuracy \citep{aghaei2021,linden2023}. It also considers nonlinear metrics, for instance, the F1 score \citep{Subramanian2023,linden2023,demirovic2021optimal,{Lin2020}}. We demonstrate in this section how our method can seamlessly accommodate these objectives within a unified framework.

An important special classification problem is  binary classification problem where $|\mathcal{K}|=2$. In this case, we denote an instance $i$ as \emph{positive} if $y_i=1$ and \emph{negative} if $y_i=0$. Here, let $n$ be the total number of samples, $n^{+}$ be the number of positive samples, and $n^{-}$ be the number of negative samples in the training set. This classification can play a crucial role in various high-risk decision domains such as risk score and healthcare. 

Table \ref{tab:cf} shows the confusion matrix for binary classification, including key elements like TP, FN, FP, and TN. TP is the count of samples predicted as positive and truly belonging to the positive class. Conversely, FN is the number of samples incorrectly predicted as negative, yet they belong to the positive class. Similarly, FP represents the count of samples falsely predicted as positive, whereas they belong to the negative class. Lastly, TN is the number of samples correctly predicted as negative and indeed belong to the negative class.

\begin{table}[htbp]
	\centering
\caption{Confusion matrix.}
	\begin{tabular}{llll}
		\hline
 ~ &               &   \multicolumn{2}{c}{Prediction}   \\ \cline{3-4}
   &   & positive &negative  \\\hline
\multirow{2}*{Label} & positive & TP                    & FN                    \\
~                    & negative  & FP                   & TN  \\ \hline
	\end{tabular}
\label{tab:cf}
\end{table}

\subsection{Cost-Sensitive Learning}
In classical machine learning methods, the primary objective is usually to minimize misclassification errors. 
However, in practical applications, the costs associated with different types of misclassifications can vary. 
For instance, misclassifying examples of the minority class as belonging to the majority class may incur higher costs than the reverse \citep{Lomax2013}. Therefore, it's more appropriate to minimize the Misclassification Error Cost (MEC):

\begin{equation}
\mathrm{MEC}=C_\mathrm{F P} \mathrm{F P} +C_{\mathrm{F N}} \mathrm{F N}
\label{costeq}
\end{equation}
where $C_{\mathrm{FP}}$ refers to the cost of misclassifying negative class samples as positive, and $C_{\mathrm{FN}}$ represents the cost of misclassifying positive samples as negative. Selecting the values for these two parameters is critical in practical machine learning problems, as they should be based on the specific application scenario. 
For example, in some cases, the cost of misclassifying negative class samples as positive may be higher, warranting a larger value for $C_{\mathrm{FP}}$. Conversely, in other situations, $C_{\mathrm{FN}}$ may be given a higher value. Appropriately setting these costs allows the model to better align with practical requirements.

Recently, several studies have focused on learning optimal classification trees with the objective of minimizing misclassification costs \citep{linden2023, maliah2021using}. Building on this, we propose modifications to the fundamental model (\ref{cp3poip}) to ensure that classification trees are effective, in cases of imbalanced datasets. 
Let $e_{tk}$ represent the number of samples incorrectly classified as label $k$ at branch node $t$. The goal for an optimal cost-sensitive classification tree is defined as follows:

\begin{equation}
\label{obj_f}
\min \quad \frac{1}{n}({\sum\limits_{t \in \mathcal{L}}C_{\mathrm{F P}}e_{t1}} + {\sum\limits_{t \in \mathcal{L}}C_{\mathrm{FN}}e_{t0}}) +\alpha \sum\limits_{f \in \mathcal{F}}\sum\limits_{t \in \mathcal{B}} a_{tf}
\end{equation}
where $\alpha$ is the complexity parameter of the tree. 
It adjusts the balance between the tree's complexity, in terms of the number of features, and the cost of in-sample misclassification errors. 
Essentially, Expression (\ref{obj_f}) implies that a feature will be discarded if its addition reduces the model's average misclassification error cost by an amount not exceeding $\alpha$.

Furthermore, we need to adjust the constraints (\ref{cp3poip}$n$)-(\ref{cp3poip}$p$) as follows:
\begin{equation}
\label{eqetk}
\begin{array}{ll}
e_{tk} \geq N_{t}-M_{k t}-n\left(1-c_{k t}\right), & \forall t \in \mathcal{L}, k \in \mathcal{K}, \\
e_{tk} \leq N_{t}-M_{k t}, & \forall t \in \mathcal{L}, k \in \mathcal{K}, \\
e_{tk} \leq nc_{k t}, & \forall t \in \mathcal{L}, k \in \mathcal{K}, \\
e_{tk} \geq 0, & \forall t \in \mathcal{L},
\end{array}
\end{equation}

The first two constraints indicate that $e_{tk}$ equals $N_{t}-M_{k t}$ if $c_{k t}=1$.
The third constraint specifies that  $e_{tk}$ is zero when $c_{k t}=0$.
Therefore, the cost-sensitive BooleanOCT model, aimed at minimizing the cost-sensitive cost, is formulated as follows:

\begin{equation}
\label{csbroct}
  \begin{array}{lll}
    \begin{array}{lll}
    \min \quad \frac{1}{n}({\sum\limits_{t \in \mathcal{L}}C_{\mathrm{F P}}e_{t1}} + {\sum\limits_{t \in \mathcal{L}}C_{\mathrm{FN}}e_{t0}}) +\alpha \sum\limits_{f \in \mathcal{F}}\sum\limits_{t \in \mathcal{B}} a_{tf}\\
   \text{Constraints} \quad(\ref{cp3poip}a) -(\ref{cp3poip}m),\\
    
    e_{tk} \geq N_{t}-M_{k t}-n\left(1-c_{k t}\right),  \quad \forall t \in \mathcal{L}, k \in \mathcal{K}, \\
    e_{tk} \leq N_{t}-M_{k t},  \quad  \forall t \in \mathcal{L}, k \in \mathcal{K}, \\
    e_{tk} \leq nc_{k t}, \quad  \forall t \in \mathcal{L}, k \in \mathcal{K}, \\
    e_{tk} \geq 0,  \quad \forall t \in \mathcal{L},
  \end{array}
  \end{array}
\end{equation}

\subsection{Balanced Accuracy}

A common strategy for addressing imbalanced datasets is to optimize what's known as balanced accuracy. 
This metric averages the accuracy across classes, offering a more comprehensive assessment of the classifier's performance for both majority and minority classes. To the best of our knowledge, only the FlowOCT model \citep{aghaei2021} is capable of learning optimal classification trees with the objective of maximizing balanced accuracy. 
Intuitively, in a two-class scenario, balanced accuracy is calculated as the average  of the true positive and true negative rates. It is calculated by:

\begin{equation}
\text{Balanced accuracy}  =1-\frac{1}{2}\left(\frac{\mathrm{F N}}{n^{+}}+\frac{\mathrm{F P}}{n^{-}}\right). 
\end{equation}

According to the expression (\ref{eqetk}), we can use $e_{tk}$ to express FN and FP, that is:
\begin{equation}
\label{FN}
\mathrm{FN}=\sum\limits_{t \in \mathcal{L}}e_{t0}
\end{equation}
\begin{equation}
\label{FP}
\mathrm{FP}=\sum\limits_{t \in \mathcal{L}}e_{t1}
\end{equation}

Thus,  the MIP formulation for constructing the optimal classification tree with the objective of maximizing balanced accuracy can be presented as follows:

\begin{equation}
\label{maxBalancemodel}
  \begin{array}{lll}
    \begin{array}{lll}
    \max \quad 1- (\frac{1}{2n^+}\sum\limits_{t \in \mathcal{L}}e_{t0} + \frac{1}{2n^-}\sum\limits_{t \in \mathcal{L}}e_{t1}) 
    \\-\alpha \sum\limits_{f \in F}\sum\limits_{t \in B} a_{tf} \\

    \text{Constraints} \quad(\ref{cp3poip}a) -(\ref{cp3poip}m),\\
    
    e_{tk} \geq N_{t}-M_{k t}-n\left(1-c_{k t}\right),  \quad \forall t \in \mathcal{L}, k \in \mathcal{K}, \\
    e_{tk} \leq N_{t}-M_{k t},  \quad  \forall t \in \mathcal{L}, k \in \mathcal{K}, \\
    e_{tk} \leq nc_{k t}, \quad  \forall t \in \mathcal{L}, k \in \mathcal{K}, \\
    e_{tk} \geq 0,  \quad \forall t \in \mathcal{L},
  \end{array}
  \end{array}
\end{equation}

\subsection{F1-score}

In binary classification problems, the F1-score is a key metric used to evaluate classifier performance, especially in imbalanced datasets. 
It is based on the recall rates of the classifier for different classes and measures the ability of the classifier to identify the majority class and minority class by calculating the harmonic mean of the recall rates for different classes, that is:
\begin{equation}
\label{f1ex}
 \operatorname{F1-score} =\frac{2\operatorname{Precision}\cdot \operatorname{Recall}}{\operatorname{Precision} + \operatorname{Recall}} =\frac{2(n^{+}-\mathrm{FN})}{2n^+ -\mathrm{FN} + \mathrm{FP}}
 \end{equation}

Let $F_1$ be the decision variable to represent the value of F1-score in the training dataset. 
The expression (\ref{f1ex}) can be rewritten using the variables $e_{t1}$ and $e_{t0}$ as shown below:
\begin{equation}
{F_1 } =  {\frac{2(n^+-\sum\limits_{t \in \mathcal{L}}e_{t0})}{2n^+-\sum\limits_{t \in \mathcal{L}}e_{t0}+\sum\limits_{t \in \mathcal{L}}e_{t1}}}
\end{equation}

Then the formulation to learn the optimal classification tree using F1-score can be designed  as the following mixed-integer quadratically constrained programming (MIQCP) model:

\begin{equation}
\label{maxF1model}
  \begin{array}{lll}
    \begin{array}{lll}
    \max \quad F_1 -\alpha \sum\limits_{f \in F}\sum\limits_{t \in B} a_{tf} \\
\text{Constraints} \quad(\ref{cp3poip}a) -(\ref{cp3poip}m),\\
    
    e_{tk} \geq N_{t}-M_{k t}-n\left(1-c_{k t}\right),  \quad \forall t \in \mathcal{L}, k \in \mathcal{K}, \\
    e_{tk} \leq N_{t}-M_{k t},  \quad  \forall t \in \mathcal{L}, k \in \mathcal{K}, \\
    e_{tk} \leq nc_{k t}, \quad  \forall t \in \mathcal{L}, k \in \mathcal{K}, \\
    e_{tk} \geq 0,  \quad \forall t \in \mathcal{L} \\

    f_1(2n^+ + \sum\limits_{t \in L}e_{t1} -\sum\limits_{t \in L}e_{t0}) \leq 2(n^+-\sum\limits_{t \in L}e_{t0}) \\

  \end{array}
  \end{array}
\end{equation}

\section{Evaluation}
\label{sec:experiments}
The aim of this section is to assess various versions of our model BooleanOCT and conduct a comparative analysis with the current state-of-the-art approaches.

\subsection{Datasets and Computational Environment}

We assessed the performance of our \textit{BooleanOCT} algorithm on 36 real-world datasets from the UCI machine learning repository. 
In our preprocessing step, datasets containing categorical and/or continuous features underwent a transformation into binary datasets. 
This transformation was achieved using a supervised discretization algorithm based on the Minimum Description Length Principle (MDLP) by \cite{fayyad1993}, which converted each feature into a categorical feature based on its statistical significance for the class. Subsequently, we employed one-hot encoding to further binarize these features.
While acknowledging that there may be more effective discretization strategies, a detailed analysis of these approaches for optimal decision trees is reserved for future work.

\textbf{Dataset}: Our study utilized 36 datasets from previous research \citep{aghaei2021,demirovic2023}. Each dataset was divided into three parts: 50\% for training, 25\% for validation, and 25\% for testing. This split was performed randomly five times, and we report the average performance across these five trials. The code and binarized datasets are accessible at https://github.com/Tommytutu/BooleanOCT.

\textbf{Model Training}: We solved the MIP-based optimal classification tree models (OCT, FlowOCT, BooleanOCT) using Gurobi 10.1 \citep{optimization2023}, setting time limits of 5 minutes for small datasets (less than 1,000 samples), 15 minutes for medium datasets (1,000 to 5,000 samples), and 30 minutes for large datasets (over 5,000 samples). 
The CART and Random Forests (RF) algorithms were implemented in Python 3.7 using scikit-learn \citep{pedregosa2011}. 
All tests were conducted on a 13th Gen Intel(R) Core(TM) i9-13900KF at 3.00 GHz with 32 GB of memory.

\subsection{Accuracy}
\label{secacc}
Previous studies have shown that DL8.5 \citep{aglin2020} outperforms BinOCT \citep{verwer2019}. Murtree \citep{Demirovic2022} demonstrates faster speed than DL8.5, GOSDT \citep{Lin2020}, and OSD \citep{hu2019}, indicating that when training time is limited, Murtree likely surpasses these models in terms of accuracy. \cite{Liu2022} found that bsnsing not only operates faster but also achieves comparable predictive accuracy to these optimal classification trees (DL8.5, BinOCT, GOSDT, OSD). Additionally, Blossom \citep{demirovic2023}, StreeD, and Murtree can derive optimal trees within seconds or minutes for tree depths less than 5. Consequently, we can infer that Blossom, StreeD, and Murtree outperform existing models like DL8.5, BinOCT, GOSDT, OSD, and bsnsing.

In this section, we compare our BooleanOCT model (\ref{cp3poip}) with the objective of minimizing classification errors against CART \citep{breiman1984}, OCT \citep{Bertsimas2017}, FlowOCT \citep{aghaei2021}, STreeD \citep{linden2023}, and the state-of-the-art black-box method, random forests \citep{breiman2001}.
The parameter settings for the algorithms (OCT, FlowOCT, BooleanOCT, CART, STreeD, RF) are detailed in Table \ref{paraOCT}.

\begin{table}[H]
\caption{Test accuracy of CART,  STreeD,  OCT, FlowOCT, RF and BooleanOCT on the small datasets.}
\begin{tabular}{lll}
\hline
Algorithm                   & Parameter                                           & Range                  \\\hline
CART                        & max\_depth: the depth of the tree                   & 1, 2, 3, 4                \\
\multirow{2}{*}{RF}         & max\_depth: the depth of the tree                   & 1, 2, 3, 4         \\
                            & n\_estimators: number of  trees                     & 50, 100, 150, 200                 \\
\multirow{2}{*}{StreeD}     & max\_depth: the depth of the tree                   & 1, 2, 3, 4              \\
                            & max\_num\_nodes:   maximum number of brance nodes   & 3, 5, 7, 9, 11, 13, 15, 17\\
\multirow{2}{*}{OCT}        & max\_depth: the depth of the tree                   & 1, 2, 3, 4               \\
                            & $\alpha$: the complexity parameter                      & 0.001, 0.01             \\
\multirow{2}{*}{FlowOCT}    & max\_depth: the depth of the tree                   & 1, 2, 3, 4                \\
                            & $\alpha$: the complexity parameter                      & 0.001, 0.01             \\
\multirow{3}{*}{BooleanOCT} & max\_depth: the depth of the tree                   & 1, 2, 3, 4                \\
                            & $\alpha$: the complexity parameter                      & 0.001, 0.01             \\
                            & $N$:   maximum number of features at the brance nodes & 3, 5                   \\\hline
\end{tabular}
\label{paraOCT}
\end{table}

 We present the results for small-scale, medium-scale, and large-scale datasets in Tables \ref{smallr}, \ref{smallm}, and \ref{smalll}, respectively. In these tables, the highest average accuracy values are highlighted in bold.
In this section, we aim to address the following questions:

\textbf{Q1} How does the performance of BooleanOCT compare to CART, OCT, FlowOCT, STreeD and RF?

\textbf{Q2} What is the impact of  multi-variable split on the performance of BooleanOCT?

\textbf{Q3} How does the difference between BooleanOCT and STreeD change as the sample size and depth increase?

\begin{table}[H]
\caption{Test accuracy of CART,  STreeD,  OCT, FlowOCT, RF and BooleanOCT on the small datasets.}
\begin{tabular}{lllllllllll}
\hline
instance         & $|\mathcal{I}|$   & $|\mathcal{F}|$   & $|\mathcal{K}|$ & CART                & STreeD         & OCT            & FlowOCT        & RF             & BooleanOCT        \\\hline
anneal           & 812 & 93  & 2 & 0.813          & 0.852          & 0.836          & 0.841          & 0.849          & \textbf{0.892} \\
appendicitis     & 106 & 530 & 2 & \textbf{0.815} & \textbf{0.815} & 0.802          & 0.815          & 0.802          & \textbf{0.815} \\
balance          & 625 & 20  & 3 & 0.679          & 0.718          & 0.715          & 0.726          & \textbf{0.860} & 0.830          \\
breast           & 277 & 38  & 2 & 0.690          & 0.681          & 0.686          & 0.671          & \textbf{0.729} & \textbf{0.729} \\
column\_3c       & 310 & 15  & 3 & 0.808          & 0.799          & 0.803          & 0.803          & \textbf{0.833} & \textbf{0.833} \\
derm\_bin        & 358 & 134 & 6 & 0.896          & 0.937          & 0.926          & 0.900          & \textbf{0.970} & 0.967          \\
dermatology      & 358 & 66  & 6 & 0.893          & 0.930          & 0.907          & 0.933          & \textbf{0.978} & 0.970          \\
diabetes         & 768 & 112 & 2 & 0.755          & 0.755          & 0.755          & 0.745          & 0.776          & \textbf{0.781} \\
hayes            & 132 & 15  & 3 & 0.616          & 0.737          & 0.747          & 0.707          & 0.828          & \textbf{0.859} \\
IndiansDiabetes  & 768 & 11  & 2 & 0.771          & 0.764          & 0.760          & 0.764          & \textbf{0.793} & 0.781          \\
Ionosphere       & 351 & 143 & 2 & 0.939          & 0.981          & 0.981          & 1.000          & 0.939          & \textbf{1.000} \\
iris             & 150 & 12  & 2 & 0.965          & 0.965          & 0.965          & 0.965          & \textbf{0.974} & 0.965          \\
monk1            & 124 & 15  & 2 & 0.785          & 0.957          & \textbf{1.000} & \textbf{1.000} & 0.774          & \textbf{1.000} \\
monk2            & 169 & 15  & 2 & 0.597          & 0.581          & 0.597          & 0.612          & 0.643          & \textbf{0.946} \\
monk3            & 122 & 15  & 2 & 0.935          & 0.935          & 0.946          & 0.935          & 0.935          & \textbf{0.957} \\
soybean          & 47  & 45  & 4 & 0.944          & 0.944          & \textbf{1.000} & 0.972          & 0.972          & \textbf{1.000} \\
spect            & 267 & 22  & 2 & 0.791          & 0.761          & 0.791          & 0.791          & \textbf{0.831} & 0.826          \\
tic              & 958 & 27  & 2 & 0.783          & 0.819          & 0.767          & 0.779          & 0.851          & \textbf{0.885} \\
vehicle          & 846 & 252 & 2 & 0.931          & 0.961          & 0.951          & 0.942          & 0.965          & \textbf{0.972} \\
wine             & 178 & 32  & 3 & 0.948          & 0.963          & 0.948          & 0.963          & \textbf{1.000} & 0.963          \\\hline
\multicolumn{4}{l}{average}      & 0.826          & 0.849          & 0.851          & 0.850          & 0.871          & 0.902   \\\hline      
\end{tabular}
\label{smallr}
\end{table}

\begin{table}[H]
\caption{Testing accuracy of CART, STreeD,  OCT, FlowOCT, RF and BooleanOCT on the small datasets.}
\begin{tabular}{llllllllll}
\hline
instance            & $|\mathcal{I}|$   & $|\mathcal{F}|$   & $|\mathcal{K}|$& CART  & OCT   & FlowOCT & StreeD         & RF             & BlooeanOCT     \\\hline

banknote            & 1372 & 16  & 2 & 0.897 & 0.940 & 0.939 & 0.936          & 0.938          & \textbf{0.941} \\
biodeg              & 1055 & 81  & 2 & 0.816 & 0.824 & 0.797 & 0.843          & 0.845          & \textbf{0.851} \\
car                 & 1728 & 19  & 4 & 0.791 & 0.783 & 0.799 & 0.810          & 0.800          & \textbf{0.869} \\
contraceptive       & 1473 & 21  & 3 & 0.551 & 0.565 & 0.556 & \textbf{0.573} & 0.559          & 0.569          \\
kr                  & 3196 & 73  & 2 & 0.945 & 0.945 & 0.945 & 0.951          & 0.949          & \textbf{0.973} \\
MaternalHealth      & 1014 & 23  & 3 & 0.816 & 0.833 & 0.824 & 0.822          & \textbf{0.835} & 0.833          \\
messidor            & 1151 & 24  & 2 & 0.622 & 0.617 & 0.620 & 0.635          & \textbf{0.653} & 0.649          \\
students            & 4424 & 97  & 3 & 0.728 & 0.733 & 0.721 & 0.747          & 0.734          & \textbf{0.748} \\
wireless            & 2000 & 42  & 4 & 0.913 & 0.927 & 0.958 & 0.959          & 0.954          & \textbf{0.973} \\
yeast               & 1484 & 89  & 2 & 0.718 & 0.701 & 0.716 & 0.711          & 0.715          & \textbf{0.740} \\
spambase            & 4600 & 132 & 2 & 0.873 & 0.869 & 0.871 & 0.885          & \textbf{0.908} & 0.902          \\\hline
\multicolumn{4}{l}{average}          & 0.788 & 0.794 & 0.795 & 0.807          & 0.808          & 0.823                  \\\hline
\end{tabular}
\label{smallm}
\end{table}

\begin{table}[H]
\caption{Testing accuracy of CART, STreeD,  OCT, FlowOCT, RF and BooleanOCT on the large datasets.}
\begin{tabular}{lllllllllllll}
\hline
instance   &  $|\mathcal{I}|$   & $|\mathcal{F}|$   & $|\mathcal{K}|$  & depth & CART  & OCT   & FlowOCT & StreeD         & RF             & BooleanOCT     \\\hline
Adult      & 32561 & 131 & 2 & 2     & 0.783 & 0.783 & 0.795   & 0.806          & 0.771          & \textbf{0.828} \\
Adult      & 32561 & 131 & 2 & 3     & 0.785 & 0.785 & 0.789   & 0.826          & 0.790          & \textbf{0.828} \\
compas     & 6907  & 12  & 2 & 2     & 0.631 & 0.656 & 0.656   & 0.656          & \textbf{0.663} & 0.662          \\
compas     & 6907  & 12  & 2 & 3     & 0.655 & 0.657 & 0.662   & 0.662          & \textbf{0.666} & 0.662          \\
eeg        & 14980 & 82  & 2 & 2     & 0.741 & 0.759 & 0.759   & \textbf{0.773} & \textbf{0.773} & 0.766          \\
eeg        & 14980 & 82  & 2 & 3     & 0.752 & 0.750 & 0.748   & \textbf{0.773} & 0.767          & 0.752          \\
fico       & 10459 & 17  & 2 & 2     & 0.691 & 0.702 & 0.712   & 0.712          & 0.709          & \textbf{0.713} \\
fico       & 10459 & 17  & 2 & 3     & 0.705 & 0.705 & 0.710   & 0.710          & \textbf{0.714} & 0.711          \\
pendigits  & 7494  & 216 & 2 & 2     & 0.970 & 0.981 & 0.981   & 0.981          & 0.971          & \textbf{0.995} \\
pendigits  & 7494  & 216 & 2 & 3     & 0.993 & 0.993 & 0.993   & 0.992          & 0.989          & \textbf{0.995} \\\hline
average    &       &     &   &       & 0.771 & 0.777 & 0.781   & 0.789          & 0.781          & 0.791       \\\hline  
\end{tabular}
\label{smalll}
\end{table}

To investigate if there is a significant difference between BooleanOCT and other models (CART, OCT, FlowOCT, STreeD, and RF), we applied the Friedman test and Nemenyi post-hoc test to the small and medium datasets. 
The results are detailed in Table \ref{FNtest}, with p-values less than 0.05 highlighted in bold. 
The Friedman test yielded a statistic of 69.216 and a corresponding p-value of 1.492e-13, indicating a significant difference among these models (BooleanOCT, CART, OCT, FlowOCT, STreeD, and RF). 
Subsequent analysis using the Nemenyi post-hoc test revealed that BooleanOCT is significantly more accurate than the other tree models (CART, OCT, FlowOCT, STreeD), addressing our first question (\textbf{Q1}).

\begin{table}[H]
\caption{Friedman test and Nemenyi post-hoc test on the small/medium datasets.}
\begin{tabular}{lllllll}
\hline
& \multicolumn{6}{c}{p-value of pairwise   comparison}         \\
                                                                               & Average rank & CART           & OCT            & FlowOCT        & StreeD         & RF    \\\hline
CART                                                                           & 4.677       &                &                &                &                &       \\
OCT                                                                            & 4.000       & 0.548          &                &                &                &       \\
FlowOCT                                                                        & 3.935       & 0.666          & 0.900          &                &                &       \\
StreeD                                                                         & 3.548       & 0.153          & 0.900          & 0.900          &                &       \\
RF                                                                             & 2.452       & \textbf{0.001} & \textbf{0.011} & 0.006          & 0.102          &       \\
BlooeanOCT                                                                     & 1.452       & \textbf{0.001} & \textbf{0.001} & \textbf{0.001} & \textbf{0.001} & 0.341 \\
\text{Friedman} $\chi^2$ & 69.216      &                &                &                &                &  \\\hline    
\end{tabular}
\label{FNtest}
\end{table}

Although the Nemenyi post-hoc test does not establish BooleanOCT as significantly more accurate than RF, a closer look reveals that BooleanOCT outperforms RF in 19 datasets, matches its performance in two, and underperforms in 10. Notably, on certain datasets, BooleanOCT substantially increases accuracy over RF. For instance, with the monk2 dataset, BooleanOCT achieves an absolute improvement of 30.3\% compared to RF. Moreover, it offers an average absolute improvement of 3.1\% on small-scale datasets and 1.5\% on medium-scale datasets (\textbf{Q1}). These results suggest that BooleanOCT is particularly suitable for small and medium datasets, due to its interpretability and accuracy..

\begin{figure}[H]
    \begin{minipage}[t]{0.49\linewidth}
        \centering
        \includegraphics[width=\textwidth]{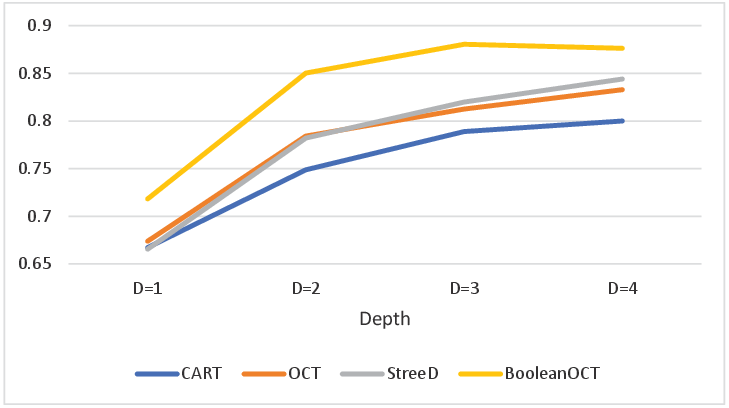}
        \centerline{(a) Small-scale datasets}
    \end{minipage}%
    \begin{minipage}[t]{0.49\linewidth}
        \centering
        \includegraphics[width=\textwidth]{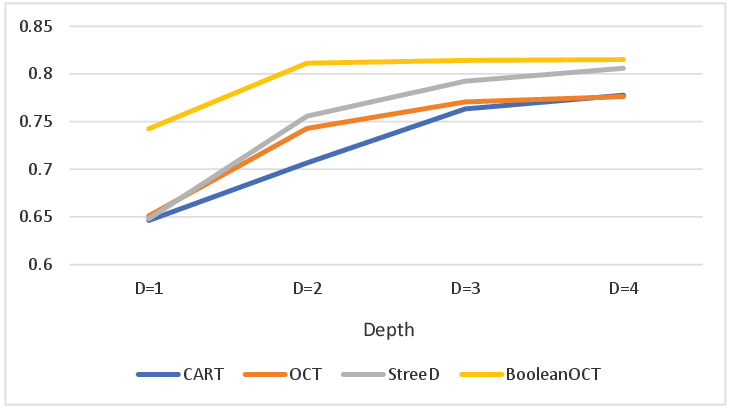}
        \centerline{(b) Medium-scale datasets}
    \end{minipage}
    \caption{Average accuracy of CART,  STreeD,  OCT, and BooleanOCT on the small/medium-scale datasets. Because OCT and Flow perform very similarly, we only present the OCT's results. The x-axis represents the depth of the tree, and the y-axis is the average accuracy.}
\label{fig:depthsm}
\end{figure}

According to Figure \ref{fig:depthsm}, the performance gap between MIP-based models (OCT and FlowOCT) and StreeD widens as tree depth increases. 
This difference is particularly noticeable in medium-scale datasets when the tree depth exceeds 2. The reason lies in the computational complexity of MIP-based models. These models, being NP-hard problems, involve approximately ${\cal O}\left( {{2^D}|\mathcal{I}|} \right)$ binary variables, where $D$ is the tree depth and $|\mathcal{I}|$ is the number of training samples. Consequently, OCT and FlowOCT struggle to derive optimal classification trees for depths greater than 2 within a 15-minute time frame for most medium-scale datasets. 
In contrast, StreeD can easily obtain optimal trees within just 10 seconds.
 As a result, as shown in Tables \ref{smallm} and \ref{smalll}, OCT and FlowOCT tend to underperform compared to StreeD on medium and large-scale datasets.

Figure \ref{fig:depthsm} (b) illustrates that the average accuracy of BooleanOCT peaks at a tree depth of 2 for medium-scale datasets. In contrast, the optimal average accuracy for other tree models (OCT, FlowOCT, StreeD, and CART) is achieved at a depth of 4. 
This suggests that for many medium-scale datasets, the number of binary decision variables in BooleanOCT (with a depth of 2) is roughly a quarter of that in OCT (with a depth of 4). 
As a result, BooleanOCT can effectively learn optimal classification trees with a shallower depth through Boolean rules, simplifying the model's complexity compared to other MIP-based models.

In summary, the Boolean rules split strategy in BooleanOCT not only facilitates better partitioning and fitting of data samples compared to single-variable split trees (CART, OCT, FlowOCT, and STreeD) but also reduces the tree depth. This reduction in depth contributes to faster solution times for BooleanOCT compared to other MIP-based models like OCT and FlowOCT (\textbf{Q1}).

MIP-based models (OCT and FlowOCT) typically achieve optimal classification trees within 5 minutes for nearly all small-scale datasets. 
 StreeD can obtain the optimal tree in mere seconds, significantly faster than the MIP-based models. Consequently, StreeD, OCT, and FlowOCT yield similar results, as shown in Table \ref{smallr}. 
 When compared to CART, these optimal classification trees (StreeD, OCT, and FlowOCT) demonstrate an average improvement of 2.4\% in accuracy, supporting the effectiveness of constructing optimal trees as established in prior studies \citep{Bertsimas2017}. 
 However, there remains a notable gap compared to RF, with RF improving average accuracy by nearly 2\%.
Applying a multi-variable split strategy (with $N=3,5$) in BooleanOCT rises its average accuracy from 0.851 (equivalent to OCT with $N=1$) to 0.902. 
The ability of BooleanOCT with multi-variable splits to create closed and open hypercubes of any dimension allows it to better fit datasets compared to single-variable split models. Therefore, we can conclude that two key factors affect accuracy: optimality and split strategy.
 Using CART as a benchmark, optimality accounts for an average absolute improvement of nearly 2.5\% (from 0.826 to 0.851), while the multi-variable split strategy contributes to an average absolute improvement of nearly 5\% (from 0.851 to 0.901), which is twice the impact of optimality. 
 This finding addresses \textbf{Q2}.

According to the Nemenyi post-hoc test, BooleanOCT significantly outperforms StreeD, but this advantage diminishes as the sample size increases, addressing \textbf{Q3}. On small-scale datasets, BooleanOCT exceeds StreeD by 5\%. This lead reduces to 1.6\% on medium-scale datasets and becomes negligible on large-scale datasets. 
This trend can be attributed to two factors. Firstly, a larger sample size enhances the generalization ability of machine learning models. Secondly, the increased sample size adds complexity to the optimal classification tree model, which may limit BooleanOCT's ability to consistently yield optimal results.
Additionally, the performance gap narrows as tree depth increases on small and medium datasets. 
This might be due to two reasons: One is that BooleanOCT achieves maximum accuracy with shallower trees compared to StreeD, as evidenced by its performance on small-scale datasets where BooleanOCT with a depth of 3 finds optimal solutions within 5 minutes for almost all cases. 
Thus, unlike StreeD, BooleanOCT does not significantly improve accuracy with increasing depth. The other reason is that BooleanOCT struggles to find optimal solutions on most medium-scale datasets when the depth exceeds 2. 
As shown in Table \ref{smalll}, StreeD, BooleanOCT, and RF have comparable accuracy, but StreeD and RF are much faster than the MIP-based models (FlowOCT, OCT, BooleanOCT).

In conclusion, we recommend using BooleanOCT for small to medium datasets due to its balance of interpretability and accuracy. For large-scale datasets, StreeD emerges as the more suitable choice, offering a combination of interpretability, faster processing speed, and comparable accuracy.

\subsection{Imbalance metrics}
\label{sec:imba}
To demonstrate the flexibility of BooleanOCT in binary classification with imbalanced datasets, we conduct experiments in three domains as introduced in Section \ref{sec:imbalanced}.

\textbf{Cost-sensitive classification}
In our cost-sensitive classification experiment, we introduce a constant misclassification cost that depends on the ratio of the two sample types.
 To be specific, if the number of positive samples exceeds  the number of negative samples,  we assign $C_{\mathrm{F N}}$ as the ratio of positive samples to negative samples and $C_{\mathrm{F P}}$ as 1. 
 Conversely, we assign $C_{\mathrm{F P}}$ as the ratio of negative samples to positive samples and $C_{\mathrm{F N}}$ as 1. The misclassification error cost  is calculated using Equation (\ref{costeq}).
 
We select two baseline algorithms, namely CSDT (Cost-Sensitive CART Decision Tree) and CSRF (Cost-Sensitive Random Forests), for this study. 
These algorithms (CSDT and CSRF) are implemented in Python 3.7 using scikit-learn \citep{pedregosa2011}, with weights assigned as $weights = {0:C_{\mathrm{F N}}, 1:C_{\mathrm{F P}}}$. 
We compare the performance of cost-sensitive BooleanOCT (CSBooleanOCT), as presented in model (\ref{csbroct}), with the two baseline algorithms (CSDT and CSRF) on 9 highly imbalanced datasets. 
The results of the cost-sensitive machine learning-based methods on these 9 datasets are summarized in Table \ref{cost_results}.

\begin{table}[H]
\centering
\caption{Misclassification error costs on the test datasets. $|\mathcal{I}|^+$ and $|\mathcal{I}|^-$ respectively represent\\ the number of positive and negative samples. The best results are bold.}
\begin{tabular}{llllllll}
\hline
instance        &  $|\mathcal{I}|$    & $|\mathcal{I}|^+$   & $|\mathcal{I}|^-$   & $|\mathcal{F}|$ & CSDT     & CSRF                & CSBooleanOCT        \\\hline
anneal          & 812  & 625  & 187  & 93  & 66.807  & 49.930           & \textbf{44.141}  \\
appendicitis    & 106  & 21   & 85   & 530 & 22.175  & 21.921           & \textbf{20.571}  \\
breast          & 277  & 81   & 196  & 38  & 36.531  & 38.284           & \textbf{33.724}  \\
IndiansDiabetes & 768  & 268  & 500  & 11  & 64.204  & \textbf{52.786}  & 58.318           \\
Ionosphere      & 351  & 25   & 326  & 143 & 43.787  & 45.120           & \textbf{0.000}   \\
spambase        & 4600 & 1913 & 2687 & 132 & 181.937 & \textbf{126.088} & 153.063          \\
spect           & 267  & 212  & 55   & 22  & 32.564  & 29.418           & \textbf{25.230}  \\
vehicle         & 846  & 218  & 628  & 252 & 21.856  & \textbf{10.174}  & 16.523           \\
yeast           & 1484 & 463  & 1021 & 89  & 165.147 & 145.130          & \textbf{139.489} \\\hline

\end{tabular}
\label{cost_results}
\end{table}

Table \ref{cost_results} presents the misclassification error costs for all methods across the nine datasets. CSBooleanOCT outperforms CSDT on all datasets, which aligns with the results in Section \ref{secacc}. This is probably due to the heuristic nature of CSDT.
When compared to the black-box model CSRF, CSRF performs better than CSDT on seven out of nine datasets. However, CSBooleanOCT also surpasses CSRF in performance in six out of nine datasets. Despite CSRF's sophisticated approach, CSBooleanOCT demonstrates competitive performance.

\textbf{Balanced accuracy} 
Several recent optimal classification models are capable of maximizing balanced accuracy \citep{Lin2020, aghaei2021}. 
Both GOSDT \citep{Lin2020} and FlowOCT \citep{aghaei2021} can obtain optimal classification trees on medium/small datasets. Hence, this study only compares BooleanOCT and FlowOCT in terms of balanced accuracy maximization. 
The out-of-sample balanced accuracy results are presented in Table \ref{fesult_Balanced}.

BooleanOCT outperforms FlowOCT on eight out of the nine datasets and matches FlowOCT's performance on one dataset. 
The p-value of the Wilcoxon signed-rank test stands at 0.012 ($<0.05$), implying a statistically significant difference in favor of BooleanOCT. BooleanOCT yields an average absolute improvement of 8\% (from 0.711 to 0.791) over FlowOCT, indicating that BooleanOCT significantly outperforms FlowOCT in terms of balanced accuracy.
This superior performance can largely be attributed to the Boolean rules split strategy employed in BooleanOCT, allowing for better partitioning and fitting of datasets compared to the single-variable split tree model (FlowOCT) as demonstrated in Section \ref{secacc}.

\begin{table}[H]
\centering
\caption{Out-of-sample  balanced accuracy. $|\mathcal{I}|^+$ and $|\mathcal{I}|^-$ respectively represent\\ the number of positive and negative samples. The best results are bold.}
\begin{tabular}{llllllll}
\hline
instance        & $|\mathcal{I}|$    & $|\mathcal{I}|^+$   & $|\mathcal{I}|^-$   & $|\mathcal{F}|$   & FolwOCT         & BooleanOCT     \\
anneal          & 812  & 625  & 187  & 93  & 0.782 & \textbf{0.870} \\
appendicitis    & 106  & 21   & 85   & 530 & 0.500 & \textbf{0.524} \\
breast          & 277  & 81   & 196  & 38  & 0.656 & 0.656          \\
IndiansDiabetes & 768  & 268  & 500  & 11  & 0.760 & \textbf{0.770} \\
Ionosphere      & 351  & 25   & 326  & 143 & 0.637 & \textbf{1.000} \\
spambase        & 4600 & 1913 & 2687 & 132 & 0.778 & \textbf{0.891} \\
spect           & 267  & 212  & 55   & 22  & 0.715 & \textbf{0.749} \\
vehicle         & 846  & 218  & 628  & 252 & 0.891 & \textbf{0.960} \\
yeast           & 1484 & 463  & 1021 & 89  & 0.677 & \textbf{0.703}\\\hline
average         &&&&  &                      0.711          &0.791\\\hline
\end{tabular}
\label{fesult_Balanced}
\end{table}

\textbf{F1 score} 
Due to the non-linear and non-convex nature of the F1-score, learning optimal trees with the objective of maximizing the F1-score can be challenging. 
A novel path-based MIP formulation was recently proposed by \cite{Subramanian2023}, leveraging column generation techniques to learn optimal multi-way split decision trees with constraints. This formulation can incorporate non-linear metrics such as F1-score. However, they did not releases their codes as open source.
Several optimal classification trees based on dynamic programming have been able to incorporate the F1-score \citep{linden2023,demirovic2021optimal,{Lin2020}}. A study by \cite{linden2023} demonstrated that StreeD performs better than both MurTree \citep{demirovic2021optimal} and GOSDT \citep{{Lin2020}}, and they have made their codes available as open source on Github.
In this study, we will therefore use StreeD as our benchmark algorithm. The average out-of-sample F1-scores obtained by StreeD and BooleanOCT are shown in Table \ref{fesult_f1}.

\begin{table}[H]
\centering
\caption{Average out-of-sample  $F_1$ score. $|\mathcal{I}|^+$ and $|\mathcal{I}|^-$ respectively represent\\ the number of positive and negative samples. The best results are bold.}
\begin{tabular}{llllllll}
\hline
instance        & $|\mathcal{I}|$    & $|\mathcal{I}|^+$   & $|\mathcal{I}|^-$   & $|\mathcal{F}|$   & StreeD         & BooleanOCT     \\
anneal          & 812  & 625  & 187  & 93  & 0.909          & \textbf{0.918} \\
appendicitis    & 106  & 21   & 85   & 530 & 0.169          & 0.169          \\
breast          & 277  & 81   & 196  & 38  & 0.490          & \textbf{0.519} \\
IndiansDiabetes & 768  & 268  & 500  & 11  & \textbf{0.688} & 0.681          \\
Ionosphere      & 351  & 25   & 326  & 143 & 0.872          & \textbf{1.000} \\
spambase        & 4600 & 1913 & 2687 & 132 & 0.872          & 0.872          \\
spect           & 267  & 212  & 55   & 22  & 0.882          & \textbf{0.894} \\
vehicle         & 846  & 218  & 628  & 252 & 0.920          & \textbf{0.932} \\
yeast           & 1484 & 463  & 1021 & 89  & 0.581          & \textbf{0.602} \\\hline
average         &&&&  &                      0.709          &0.732\\\hline
\end{tabular}
\label{fesult_f1}
\end{table}

The performance of BooleanOCT surpasses that of StreeD on 6 out of 9 datasets, and matches the performance of StreeD on 2 datasets. 
Alongside this, the p-value of the Wilcoxon signed rank test is 0.028, which is less than the standard significance level of 0.05. This indicates that the BooleanOCT's performance is statistically significantly better than StreeD's. 
Moreover, BooleanOCT provides an average absolute improvement of 2.3\% (increasing from 0.709 to 0.732) over StreeD. 
Therefore, the conclusion can be drawn that, in terms of maximizing the F1-score on these 9 datasets, the performance of BooleanOCT is significantly superior to that of StreeD.

\section{Conclusion and Future Work}
\label{sec:conclusion}

In this paper, we introduce BooleanOCT, a new MIP formulation of optimal classification trees that utilizes Boolean rules. 
The multivariate Boolean rules split strategy employed in BooleanOCT has the capability to reduce the overall partitioning of the data space. This is achieved by permitting multiple features in a branch node to capture the  heterogeneous characteristics of instances within the same subregion.
Consequently, BooleanOCT not only supports better partitioning and fitting of data samples than single-variable split trees like CART, OCT, FlowOCT, and STreeD, but it also reduces the tree depth, resulting in faster solving speed than MIP-based tree models like OCT and FlowOCT. 
Our experiments show that BooleanOCT significantly outperforms other optimal classification tree models such as OCT, FlowOCT, and StreeD. For small-scale and medium-sized datasets, BooleanOCT improves  the average accuracy of RF by 3.1\% and 1.5\%, respectively.

Moreover, BooleanOCT can handle binary classification problems with imbalanced datasets by incorporating linear metrics like balanced accuracy and nonlinear metrics like F1-score. Our experimental results, detailed in Section \ref{sec:imba}, demonstrate that BooleanOCT is comparable to or better than the contemporary methods in the literature for optimal classification trees. 
Due to its interpretability and accuracy, BooleanOCT is particularly promising for high-risk decision-making sectors such as credit scoring and healthcare.

Future work will focus on several areas. Since the BooleanOCT proposed in this paper is an NP-hard problem, applying this model to large-scale datasets could be a challenge. We plan to use more efficient algorithms like the Bender's decomposition \citep{aghaei2021} and column generation \citep{Subramanian2023} to tackle this. Furthermore, since optimal classification trees based on multi-way split can fit datasets better than those based on two-way split, we plan to enhance the performance of BooleanOCT by introducing multi-way split.


\section*{Acknowledgment}
{This work was supported by the National Natural Science Foundation of China under Grants 72371175 and 71971148 and by the Fundamental Research Funds for the Central Universities under Grants 2023ZY-SX020, 2023ZY-SX023, and SXYPY202334.}	
		
		

\begingroup
\setlength{\bibsep}{0pt}
\bibliographystyle{elsarticle-harv}
\bibliography{fodt}

\begin{thebibliography}{62}
\expandafter\ifx\csname natexlab\endcsname\relax\def\natexlab#1{#1}\fi
\providecommand{\url}[1]{\texttt{#1}}
\providecommand{\href}[2]{#2}
\providecommand{\path}[1]{#1}
\providecommand{\DOIprefix}{doi:}
\providecommand{\ArXivprefix}{arXiv:}
\providecommand{\URLprefix}{URL: }
\providecommand{\Pubmedprefix}{pmid:}
\providecommand{\doi}[1]{\href{http://dx.doi.org/#1}{\path{#1}}}
\providecommand{\Pubmed}[1]{\href{pmid:#1}{\path{#1}}}
\providecommand{\bibinfo}[2]{#2}
\ifx\xfnm\relax \def\xfnm[#1]{\unskip,\space#1}\fi
\bibitem[{Aghaei et~al.(2021)Aghaei, G{\'o}mez and Vayanos}]{aghaei2021}
\bibinfo{author}{Aghaei, S.}, \bibinfo{author}{G{\'o}mez, A.},
  \bibinfo{author}{Vayanos, P.}, \bibinfo{year}{2021}.
\newblock \bibinfo{title}{Strong optimal classification trees}.
\newblock \bibinfo{journal}{arXiv preprint arXiv:2103.15965} .
\bibitem[{Aglin et~al.(2020)Aglin, Nijssen and Schaus}]{aglin2020}
\bibinfo{author}{Aglin, G.}, \bibinfo{author}{Nijssen, S.},
  \bibinfo{author}{Schaus, P.}, \bibinfo{year}{2020}.
\newblock \bibinfo{title}{Learning optimal decision trees using caching
  branch-and-bound search}, in: \bibinfo{booktitle}{Proceedings of the AAAI
  conference on Artificial Intelligence}, pp. \bibinfo{pages}{3146--3153}.
\bibitem[{Angelino et~al.(2018)Angelino, Larus-Stone, Alabi, Seltzer and
  Rudin}]{angelino2018learning}
\bibinfo{author}{Angelino, E.}, \bibinfo{author}{Larus-Stone, N.},
  \bibinfo{author}{Alabi, D.}, \bibinfo{author}{Seltzer, M.},
  \bibinfo{author}{Rudin, C.}, \bibinfo{year}{2018}.
\newblock \bibinfo{title}{Learning certifiably optimal rule lists for
  categorical data}.
\newblock \bibinfo{journal}{Journal of Machine Learning Research}
  \bibinfo{volume}{18}, \bibinfo{pages}{1--78}.
\bibitem[{Aouad et~al.(2023)Aouad, Elmachtoub, Ferreira and
  McNellis}]{aouad2023}
\bibinfo{author}{Aouad, A.}, \bibinfo{author}{Elmachtoub, A.N.},
  \bibinfo{author}{Ferreira, K.J.}, \bibinfo{author}{McNellis, R.},
  \bibinfo{year}{2023}.
\newblock \bibinfo{title}{Market segmentation trees}.
\newblock \bibinfo{journal}{Manufacturing \& Service Operations Management}
  \bibinfo{volume}{25}, \bibinfo{pages}{648--667}.
\bibitem[{Avellaneda(2020)}]{avellaneda2020efficient}
\bibinfo{author}{Avellaneda, F.}, \bibinfo{year}{2020}.
\newblock \bibinfo{title}{Efficient inference of optimal decision trees}, in:
  \bibinfo{booktitle}{Proceedings of the AAAI conference on Artificial
  Intelligence}, pp. \bibinfo{pages}{3195--3202}.
\bibitem[{Bagallo and Haussler(1990)}]{bagallo1990}
\bibinfo{author}{Bagallo, G.}, \bibinfo{author}{Haussler, D.},
  \bibinfo{year}{1990}.
\newblock \bibinfo{title}{Boolean feature discovery in empirical learning}.
\newblock \bibinfo{journal}{Machine Learning} \bibinfo{volume}{5},
  \bibinfo{pages}{71--99}.
\bibitem[{Balvert(2024)}]{Balvert2024}
\bibinfo{author}{Balvert, M.}, \bibinfo{year}{2024}.
\newblock \bibinfo{title}{Iterative rule extension for logic analysis of data:
  An milp-based heuristic to derive interpretable binary classifiers from large
  data sets}.
\newblock \bibinfo{journal}{INFORMS Journal on Computing}
  \DOIprefix\doi{10.1287/ijoc.2021.0284}.
\bibitem[{Bertsimas and Dunn(2017)}]{Bertsimas2017}
\bibinfo{author}{Bertsimas, D.}, \bibinfo{author}{Dunn, J.},
  \bibinfo{year}{2017}.
\newblock \bibinfo{title}{{Optimal classification trees}}.
\newblock \bibinfo{journal}{Machine Learning} \bibinfo{volume}{106},
  \bibinfo{pages}{1039--1082}.
\bibitem[{Bertsimas et~al.(2022)Bertsimas, Dunn and
  Paskov}]{bertsimas2022stable}
\bibinfo{author}{Bertsimas, D.}, \bibinfo{author}{Dunn, J.},
  \bibinfo{author}{Paskov, I.}, \bibinfo{year}{2022}.
\newblock \bibinfo{title}{Stable classification}.
\newblock \bibinfo{journal}{Journal of Machine Learning Research}
  \bibinfo{volume}{23}, \bibinfo{pages}{13401--13453}.
\bibitem[{Bertsimas et~al.(2019)Bertsimas, Dunn, Pawlowski and
  Zhuo}]{bertsimas2019}
\bibinfo{author}{Bertsimas, D.}, \bibinfo{author}{Dunn, J.},
  \bibinfo{author}{Pawlowski, C.}, \bibinfo{author}{Zhuo, Y.D.},
  \bibinfo{year}{2019}.
\newblock \bibinfo{title}{Robust classification}.
\newblock \bibinfo{journal}{INFORMS Journal on Optimization}
  \bibinfo{volume}{1}, \bibinfo{pages}{2--34}.
\bibitem[{Bertsimas and Kallus(2020)}]{bertsimas2020}
\bibinfo{author}{Bertsimas, D.}, \bibinfo{author}{Kallus, N.},
  \bibinfo{year}{2020}.
\newblock \bibinfo{title}{From predictive to prescriptive analytics}.
\newblock \bibinfo{journal}{Management Science} \bibinfo{volume}{66},
  \bibinfo{pages}{1025--1044}.
\bibitem[{Bertsimas and Shioda(2007)}]{bertsimas2007}
\bibinfo{author}{Bertsimas, D.}, \bibinfo{author}{Shioda, R.},
  \bibinfo{year}{2007}.
\newblock \bibinfo{title}{Classification and regression via integer
  optimization}.
\newblock \bibinfo{journal}{Operations research} \bibinfo{volume}{55},
  \bibinfo{pages}{252--271}.
\bibitem[{Blanco et~al.(2023)Blanco, Jap{\'o}n and Puerto}]{blanco2023}
\bibinfo{author}{Blanco, V.}, \bibinfo{author}{Jap{\'o}n, A.},
  \bibinfo{author}{Puerto, J.}, \bibinfo{year}{2023}.
\newblock \bibinfo{title}{Multiclass optimal classification trees with
  svm-splits}.
\newblock \bibinfo{journal}{Machine Learning} \bibinfo{volume}{112},
  \bibinfo{pages}{4905--4928}.
\bibitem[{Boutilier et~al.(2023)Boutilier, Michini and
  Zhou}]{boutilier2023optimal}
\bibinfo{author}{Boutilier, J.}, \bibinfo{author}{Michini, C.},
  \bibinfo{author}{Zhou, Z.}, \bibinfo{year}{2023}.
\newblock \bibinfo{title}{Optimal multivariate decision trees}.
\newblock \bibinfo{journal}{Constraints} , \bibinfo{pages}{1--29}.
\bibitem[{Breiman(2001)}]{breiman2001}
\bibinfo{author}{Breiman, L.}, \bibinfo{year}{2001}.
\newblock \bibinfo{title}{Random forests}.
\newblock \bibinfo{journal}{Machine learning} \bibinfo{volume}{45},
  \bibinfo{pages}{5--32}.
\bibitem[{Breiman et~al.(1984)Breiman, Friedman, Olshen and
  Stone}]{breiman1984}
\bibinfo{author}{Breiman, L.}, \bibinfo{author}{Friedman, J.},
  \bibinfo{author}{Olshen, R.}, \bibinfo{author}{Stone, C.},
  \bibinfo{year}{1984}.
\newblock \bibinfo{title}{Classification and regression trees}.
\newblock \bibinfo{publisher}{Monterey, CA: Wadsworth and Brooks}.
\bibitem[{Carrizosa et~al.(2021)Carrizosa, Molero-Rio and
  Romero~Morales}]{carrizosa2021}
\bibinfo{author}{Carrizosa, E.}, \bibinfo{author}{Molero-Rio, C.},
  \bibinfo{author}{Romero~Morales, D.}, \bibinfo{year}{2021}.
\newblock \bibinfo{title}{Mathematical optimization in classification and
  regression trees}.
\newblock \bibinfo{journal}{Top} \bibinfo{volume}{29}, \bibinfo{pages}{5--33}.
\bibitem[{Chen et~al.(2022)Chen, Lin, Rudin, Shaposhnik, Wang and
  Wang}]{Chen2022}
\bibinfo{author}{Chen, C.}, \bibinfo{author}{Lin, K.}, \bibinfo{author}{Rudin,
  C.}, \bibinfo{author}{Shaposhnik, Y.}, \bibinfo{author}{Wang, S.},
  \bibinfo{author}{Wang, T.}, \bibinfo{year}{2022}.
\newblock \bibinfo{title}{A holistic approach to interpretability in financial
  lending: Models, visualizations, and summary-explanations}.
\newblock \bibinfo{journal}{Decision Support Systems} \bibinfo{volume}{152},
  \bibinfo{pages}{113647}.
\bibitem[{Ciocan and Mi{\v{s}}i{\'c}(2022)}]{ciocan2022}
\bibinfo{author}{Ciocan, D.F.}, \bibinfo{author}{Mi{\v{s}}i{\'c}, V.V.},
  \bibinfo{year}{2022}.
\newblock \bibinfo{title}{Interpretable optimal stopping}.
\newblock \bibinfo{journal}{Management Science} \bibinfo{volume}{68},
  \bibinfo{pages}{1616--1638}.
\bibitem[{Costa and Pedreira(2023)}]{costa2023recent}
\bibinfo{author}{Costa, V.G.}, \bibinfo{author}{Pedreira, C.E.},
  \bibinfo{year}{2023}.
\newblock \bibinfo{title}{Recent advances in decision trees: An updated
  survey}.
\newblock \bibinfo{journal}{Artificial Intelligence Review}
  \bibinfo{volume}{56}, \bibinfo{pages}{4765--4800}.
\bibitem[{Demirovi{\'c} et~al.(2023)Demirovi{\'c}, Hebrard and
  Jean}]{demirovic2023}
\bibinfo{author}{Demirovi{\'c}, E.}, \bibinfo{author}{Hebrard, E.},
  \bibinfo{author}{Jean, L.}, \bibinfo{year}{2023}.
\newblock \bibinfo{title}{Blossom: an anytime algorithm for computing optimal
  decision trees}, in: \bibinfo{booktitle}{International Conference on Machine
  Learning}, \bibinfo{organization}{JMLR. org}.
\bibitem[{Demirovi{\'{c}} et~al.(2022)Demirovi{\'{c}}, Lukina, Hebrard, Chan,
  Bailey, Leckie, Ramamohanarao and Stuckey}]{Demirovic2022}
\bibinfo{author}{Demirovi{\'{c}}, E.}, \bibinfo{author}{Lukina, A.},
  \bibinfo{author}{Hebrard, E.}, \bibinfo{author}{Chan, J.},
  \bibinfo{author}{Bailey, J.}, \bibinfo{author}{Leckie, C.},
  \bibinfo{author}{Ramamohanarao, K.}, \bibinfo{author}{Stuckey, P.J.},
  \bibinfo{year}{2022}.
\newblock \bibinfo{title}{Murtree: Optimal decision trees via dynamic
  programming and search}.
\newblock \bibinfo{journal}{Journal of Machine Learning Research}
  \bibinfo{volume}{23}, \bibinfo{pages}{1--47}.
\bibitem[{Demirovi{\'c} and Stuckey(2021)}]{demirovic2021optimal}
\bibinfo{author}{Demirovi{\'c}, E.}, \bibinfo{author}{Stuckey, P.J.},
  \bibinfo{year}{2021}.
\newblock \bibinfo{title}{Optimal decision trees for nonlinear metrics}, in:
  \bibinfo{booktitle}{Proceedings of the AAAI conference on Artificial
  Intelligence}, pp. \bibinfo{pages}{3733--3741}.
\bibitem[{Dunn(2018)}]{dunn2018optimal}
\bibinfo{author}{Dunn, J.W.}, \bibinfo{year}{2018}.
\newblock \bibinfo{title}{Optimal trees for prediction and prescription}.
\newblock Ph.D. thesis. Massachusetts Institute of Technology.
\bibitem[{Elmachtoub et~al.(2020)Elmachtoub, Liang and
  McNellis}]{elmachtoub2020}
\bibinfo{author}{Elmachtoub, A.N.}, \bibinfo{author}{Liang, J.C.N.},
  \bibinfo{author}{McNellis, R.}, \bibinfo{year}{2020}.
\newblock \bibinfo{title}{Decision trees for decision-making under the
  predict-then-optimize framework}, in: \bibinfo{booktitle}{International
  Conference on Machine Learning}, \bibinfo{organization}{PMLR}. pp.
  \bibinfo{pages}{2858--2867}.
\bibitem[{Fayyad and Irani(1993)}]{fayyad1993}
\bibinfo{author}{Fayyad, U.M.}, \bibinfo{author}{Irani, K.B.},
  \bibinfo{year}{1993}.
\newblock \bibinfo{title}{Multi-interval discretization of continuous-valued
  attributes for classification learning}, in:
  \bibinfo{booktitle}{International Joint Conferences on Artificial
  Intelligence}, \bibinfo{organization}{Citeseer}. pp.
  \bibinfo{pages}{1022--1029}.
\bibitem[{Firat et~al.(2020)Firat, Crognier, Gabor, Hurkens and
  Zhang}]{firat2020}
\bibinfo{author}{Firat, M.}, \bibinfo{author}{Crognier, G.},
  \bibinfo{author}{Gabor, A.F.}, \bibinfo{author}{Hurkens, C.A.},
  \bibinfo{author}{Zhang, Y.}, \bibinfo{year}{2020}.
\newblock \bibinfo{title}{Column generation based heuristic for learning
  classification trees}.
\newblock \bibinfo{journal}{Computers \& Operations Research}
  \bibinfo{volume}{116}, \bibinfo{pages}{104866}.
\bibitem[{G{\"u}nl{\"u}k et~al.(2021)G{\"u}nl{\"u}k, Kalagnanam, Li, Menickelly
  and Scheinberg}]{gunluk2021}
\bibinfo{author}{G{\"u}nl{\"u}k, O.}, \bibinfo{author}{Kalagnanam, J.},
  \bibinfo{author}{Li, M.}, \bibinfo{author}{Menickelly, M.},
  \bibinfo{author}{Scheinberg, K.}, \bibinfo{year}{2021}.
\newblock \bibinfo{title}{Optimal decision trees for categorical data via
  integer programming}.
\newblock \bibinfo{journal}{Journal of Global Optimization}
  \bibinfo{volume}{81}, \bibinfo{pages}{233--260}.
\bibitem[{Gurobi~Optimization(2023)}]{optimization2023}
\bibinfo{author}{Gurobi~Optimization, L.}, \bibinfo{year}{2023}.
\newblock \bibinfo{title}{Gurobi optimizer reference manual}.
\newblock \bibinfo{journal}{https://www.gurobi.com/solutions/gurobi-optimizer}
  , \bibinfo{pages}{Accessed date: 9 October 2023}.
\bibitem[{Hu et~al.(2019)Hu, Rudin and Seltzer}]{hu2019}
\bibinfo{author}{Hu, X.}, \bibinfo{author}{Rudin, C.},
  \bibinfo{author}{Seltzer, M.}, \bibinfo{year}{2019}.
\newblock \bibinfo{title}{Optimal sparse decision trees}.
\newblock \bibinfo{journal}{Advances in Neural Information Processing Systems}
  \bibinfo{volume}{32}.
\bibitem[{Hua et~al.(2022)Hua, Ren and Cao}]{hua2022scalable}
\bibinfo{author}{Hua, K.}, \bibinfo{author}{Ren, J.}, \bibinfo{author}{Cao,
  Y.}, \bibinfo{year}{2022}.
\newblock \bibinfo{title}{A scalable deterministic global optimization
  algorithm for training optimal decision tree}.
\newblock \bibinfo{journal}{Advances in Neural Information Processing Systems}
  \bibinfo{volume}{35}, \bibinfo{pages}{8347--8359}.
\bibitem[{Laugel et~al.(2019)Laugel, Lesot, Marsala, Renard and
  Detyniecki}]{laugel201911}
\bibinfo{author}{Laugel, T.}, \bibinfo{author}{Lesot, M.J.},
  \bibinfo{author}{Marsala, C.}, \bibinfo{author}{Renard, X.},
  \bibinfo{author}{Detyniecki, M.}, \bibinfo{year}{2019}.
\newblock \bibinfo{title}{The dangers of post-hoc interpretability: Unjustified
  counterfactual explanations}, in: \bibinfo{booktitle}{Proceedings of the
  Twenty-Eighth International Joint Conference on Artificial Intelligence}, pp.
  \bibinfo{pages}{2801--2807}.
\bibitem[{Laurent and Rivest(1976)}]{laurent1976}
\bibinfo{author}{Laurent, H.}, \bibinfo{author}{Rivest, R.L.},
  \bibinfo{year}{1976}.
\newblock \bibinfo{title}{{Constructing optimal binary decision trees is
  NP-complete}}.
\newblock \bibinfo{journal}{Information Processing Letters}
  \bibinfo{volume}{5}, \bibinfo{pages}{15--17}.
\bibitem[{Lin et~al.(2020)Lin, Zhong, Hu, Rudin and Seltzer}]{Lin2020}
\bibinfo{author}{Lin, J.}, \bibinfo{author}{Zhong, C.}, \bibinfo{author}{Hu,
  D.}, \bibinfo{author}{Rudin, C.}, \bibinfo{author}{Seltzer, M.},
  \bibinfo{year}{2020}.
\newblock \bibinfo{title}{Generalized and scalable optimal sparse decision
  trees}, in: \bibinfo{booktitle}{International Conference on Machine
  Learning}, pp. \bibinfo{pages}{6150--6160}.
\bibitem[{van~der Linden et~al.(2023)van~der Linden, de~Weerdt and
  Demirovi{\'c}}]{linden2023}
\bibinfo{author}{van~der Linden, J.G.}, \bibinfo{author}{de~Weerdt, M.},
  \bibinfo{author}{Demirovi{\'c}, E.}, \bibinfo{year}{2023}.
\newblock \bibinfo{title}{Necessary and sufficient conditions for optimal
  decision trees using dynamic programming}, in:
  \bibinfo{booktitle}{Thirty-seventh Conference on Neural Information
  Processing Systems}.
\bibitem[{Liu(2022)}]{Liu2022}
\bibinfo{author}{Liu, Y.}, \bibinfo{year}{2022}.
\newblock \bibinfo{title}{bsnsing: A decision tree induction method based on
  recursive optimal boolean rule composition}.
\newblock \bibinfo{journal}{INFORMS Journal on Computing} \bibinfo{volume}{34},
  \bibinfo{pages}{2908--2929}.
\bibitem[{Lomax and Vadera(2013)}]{Lomax2013}
\bibinfo{author}{Lomax, S.}, \bibinfo{author}{Vadera, S.},
  \bibinfo{year}{2013}.
\newblock \bibinfo{title}{{A survey of cost-sensitive decision tree induction
  algorithms}}.
\newblock \bibinfo{journal}{ACM Computing Surveys} \bibinfo{volume}{45},
  \bibinfo{pages}{1--35}.
\bibitem[{Lou et~al.(2012)Lou, Caruana and Gehrke}]{Lou2012}
\bibinfo{author}{Lou, Y.}, \bibinfo{author}{Caruana, R.},
  \bibinfo{author}{Gehrke, J.}, \bibinfo{year}{2012}.
\newblock \bibinfo{title}{Intelligible models for classification and
  regression}, in: \bibinfo{booktitle}{Proceedings of the 18th ACM SIGKDD
  International Conference on Knowledge Discovery and Data Mining}, pp.
  \bibinfo{pages}{150--158}.
\bibitem[{Lundberg and Lee(2017)}]{Lundberg2017}
\bibinfo{author}{Lundberg, S.M.}, \bibinfo{author}{Lee, S.I.},
  \bibinfo{year}{2017}.
\newblock \bibinfo{title}{A unified approach to interpreting model
  predictions}.
\newblock \bibinfo{journal}{Advances in Neural Information Processing Systems}
  \bibinfo{volume}{30}.
\bibitem[{Maliah and Shani(2021)}]{maliah2021using}
\bibinfo{author}{Maliah, S.}, \bibinfo{author}{Shani, G.},
  \bibinfo{year}{2021}.
\newblock \bibinfo{title}{Using pomdps for learning cost sensitive decision
  trees}.
\newblock \bibinfo{journal}{Artificial intelligence} \bibinfo{volume}{292},
  \bibinfo{pages}{103400}.
\bibitem[{Mazumder et~al.(2022)Mazumder, Meng and Wang}]{mazumder2022quant}
\bibinfo{author}{Mazumder, R.}, \bibinfo{author}{Meng, X.},
  \bibinfo{author}{Wang, H.}, \bibinfo{year}{2022}.
\newblock \bibinfo{title}{Quant-bnb: A scalable branch-and-bound method for
  optimal decision trees with continuous features}, in:
  \bibinfo{booktitle}{International Conference on Machine Learning}, pp.
  \bibinfo{pages}{15255--15277}.
\bibitem[{McTavish et~al.(2022)McTavish, Zhong, Achermann, Karimalis, Chen,
  Rudin and Seltzer}]{mctavish2022fast}
\bibinfo{author}{McTavish, H.}, \bibinfo{author}{Zhong, C.},
  \bibinfo{author}{Achermann, R.}, \bibinfo{author}{Karimalis, I.},
  \bibinfo{author}{Chen, J.}, \bibinfo{author}{Rudin, C.},
  \bibinfo{author}{Seltzer, M.}, \bibinfo{year}{2022}.
\newblock \bibinfo{title}{Fast sparse decision tree optimization via reference
  ensembles}, in: \bibinfo{booktitle}{Proceedings of the AAAI Conference on
  Artificial Intelligence}, pp. \bibinfo{pages}{9604--9613}.
\bibitem[{Narodytska et~al.(2018)Narodytska, Ignatiev, Pereira and
  Marques-Silva}]{Nina2018}
\bibinfo{author}{Narodytska, N.}, \bibinfo{author}{Ignatiev, A.},
  \bibinfo{author}{Pereira, F.}, \bibinfo{author}{Marques-Silva, J.},
  \bibinfo{year}{2018}.
\newblock \bibinfo{title}{Learning optimal decision trees with {SAT}}, in:
  \bibinfo{booktitle}{Proceedings of the Twenty-Seventh International Joint
  Conference on Artificial Intelligence, {IJCAI-18}}, pp.
  \bibinfo{pages}{1362--1368}.
\bibitem[{Nijssen and Fromont(2007)}]{nijssen2007mining}
\bibinfo{author}{Nijssen, S.}, \bibinfo{author}{Fromont, E.},
  \bibinfo{year}{2007}.
\newblock \bibinfo{title}{Mining optimal decision trees from itemset lattices},
  in: \bibinfo{booktitle}{Proceedings of the 13th ACM SIGKDD International
  Conference on Knowledge Discovery and Data Mining}, pp.
  \bibinfo{pages}{530--539}.
\bibitem[{Nijssen and Fromont(2010)}]{nijssen2010optimal}
\bibinfo{author}{Nijssen, S.}, \bibinfo{author}{Fromont, E.},
  \bibinfo{year}{2010}.
\newblock \bibinfo{title}{Optimal constraint-based decision tree induction from
  itemset lattices}.
\newblock \bibinfo{journal}{Data Mining and Knowledge Discovery}
  \bibinfo{volume}{21}, \bibinfo{pages}{9--51}.
\bibitem[{{Parliament and Council of the European
  Union}(2016)}]{Parliament2016}
\bibinfo{author}{{Parliament and Council of the European Union}},
  \bibinfo{year}{2016}.
\newblock \bibinfo{title}{General data protection regulation}.
\newblock \bibinfo{journal}{https://gdpr-info.eu/chapter-3/} ,
  \bibinfo{pages}{Accessed date: 9 October 2023}.
\bibitem[{Pedregosa et~al.(2011)Pedregosa, Varoquaux, Gramfort, Michel,
  Thirion, Grisel, Blondel, Prettenhofer, Weiss, Dubourg
  et~al.}]{pedregosa2011}
\bibinfo{author}{Pedregosa, F.}, \bibinfo{author}{Varoquaux, G.},
  \bibinfo{author}{Gramfort, A.}, \bibinfo{author}{Michel, V.},
  \bibinfo{author}{Thirion, B.}, \bibinfo{author}{Grisel, O.},
  \bibinfo{author}{Blondel, M.}, \bibinfo{author}{Prettenhofer, P.},
  \bibinfo{author}{Weiss, R.}, \bibinfo{author}{Dubourg, V.}, et~al.,
  \bibinfo{year}{2011}.
\newblock \bibinfo{title}{Scikit-learn: Machine learning in python}.
\newblock \bibinfo{journal}{the Journal of machine Learning research}
  \bibinfo{volume}{12}, \bibinfo{pages}{2825--2830}.
\bibitem[{Quinlan(1986)}]{quinlan1986}
\bibinfo{author}{Quinlan, J.R.}, \bibinfo{year}{1986}.
\newblock \bibinfo{title}{Induction of decision trees}.
\newblock \bibinfo{journal}{Machine Learning} \bibinfo{volume}{1},
  \bibinfo{pages}{81--106}.
\bibitem[{Quinlan(1993)}]{quinlan1993}
\bibinfo{author}{Quinlan, J.R.}, \bibinfo{year}{1993}.
\newblock \bibinfo{title}{C4. 5: programs for machine learning}.
\newblock \bibinfo{publisher}{Morgan Kaufmann, San Francisco, CA}.
\bibitem[{Ragodos and Wang(2022)}]{ragodos2022}
\bibinfo{author}{Ragodos, R.}, \bibinfo{author}{Wang, T.},
  \bibinfo{year}{2022}.
\newblock \bibinfo{title}{Disjunctive rule lists}.
\newblock \bibinfo{journal}{INFORMS Journal On Computing} \bibinfo{volume}{34},
  \bibinfo{pages}{3259--3276}.
\bibitem[{Ribeiro et~al.(2016)Ribeiro, Singh and Guestrin}]{Ribeiro2016}
\bibinfo{author}{Ribeiro, M.T.}, \bibinfo{author}{Singh, S.},
  \bibinfo{author}{Guestrin, C.}, \bibinfo{year}{2016}.
\newblock \bibinfo{title}{{``Why should I trust you?"}: Explaining the
  predictions of any classifier}, in: \bibinfo{booktitle}{Proceedings of the
  22nd ACM SIGKDD International Conference on Knowledge Discovery and Data
  Mining}, pp. \bibinfo{pages}{1135--1144}.
\bibitem[{Ross et~al.(2017)Ross, Hughes and Doshi-Velez}]{ross2017}
\bibinfo{author}{Ross, A.S.}, \bibinfo{author}{Hughes, M.C.},
  \bibinfo{author}{Doshi-Velez, F.}, \bibinfo{year}{2017}.
\newblock \bibinfo{title}{Right for the right reasons: Training differentiable
  models by constraining their explanations}, in:
  \bibinfo{booktitle}{Proceedings of the Twenty-Sixth International Joint
  Conference on Artificial Intelligence, {IJCAI-17}}, pp.
  \bibinfo{pages}{2662--2670}.
\bibitem[{Rudin(2019)}]{rudin2019}
\bibinfo{author}{Rudin, C.}, \bibinfo{year}{2019}.
\newblock \bibinfo{title}{{Stop explaining black box machine learning models
  for high stakes decisions and use interpretable models instead}}.
\newblock \bibinfo{journal}{Nature Machine Intelligence} \bibinfo{volume}{1},
  \bibinfo{pages}{206--215}.
\bibitem[{Shati et~al.(2023)Shati, Cohen and McIlraith}]{Shati2023}
\bibinfo{author}{Shati, P.}, \bibinfo{author}{Cohen, E.},
  \bibinfo{author}{McIlraith, S.A.}, \bibinfo{year}{2023}.
\newblock \bibinfo{title}{{SAT-based optimal classification trees for
  non-binary data}}.
\newblock \bibinfo{journal}{Constraints} \bibinfo{volume}{28},
  \bibinfo{pages}{166--202}.
\bibitem[{Slack et~al.(2020)Slack, Hilgard, Jia, Singh and
  Lakkaraju}]{Slack2020}
\bibinfo{author}{Slack, D.}, \bibinfo{author}{Hilgard, S.},
  \bibinfo{author}{Jia, E.}, \bibinfo{author}{Singh, S.},
  \bibinfo{author}{Lakkaraju, H.}, \bibinfo{year}{2020}.
\newblock \bibinfo{title}{Fooling lime and shap: Adversarial attacks on post
  hoc explanation methods}, in: \bibinfo{booktitle}{Proceedings of the AAAI/ACM
  Conference on AI, Ethics, and Society}, pp. \bibinfo{pages}{180--186}.
\bibitem[{Subramanian and Sun(2023)}]{Subramanian2023}
\bibinfo{author}{Subramanian, S.}, \bibinfo{author}{Sun, W.},
  \bibinfo{year}{2023}.
\newblock \bibinfo{title}{Scalable optimal multiway-split decision trees with
  constraints}.
\newblock \bibinfo{journal}{Proceedings of the AAAI Conference on Artificial
  Intelligence} \bibinfo{volume}{37}, \bibinfo{pages}{9891--9899}.
\bibitem[{Ustun and Rudin(2016)}]{ustun2016}
\bibinfo{author}{Ustun, B.}, \bibinfo{author}{Rudin, C.}, \bibinfo{year}{2016}.
\newblock \bibinfo{title}{Supersparse linear integer models for optimized
  medical scoring systems}.
\newblock \bibinfo{journal}{Machine Learning} \bibinfo{volume}{102},
  \bibinfo{pages}{349--391}.
\bibitem[{Verhaeghe et~al.(2020)Verhaeghe, Nijssen, Pesant, Quimper and
  Schaus}]{verhaeghe2020learning}
\bibinfo{author}{Verhaeghe, H.}, \bibinfo{author}{Nijssen, S.},
  \bibinfo{author}{Pesant, G.}, \bibinfo{author}{Quimper, C.G.},
  \bibinfo{author}{Schaus, P.}, \bibinfo{year}{2020}.
\newblock \bibinfo{title}{Learning optimal decision trees using constraint
  programming}.
\newblock \bibinfo{journal}{Constraints} \bibinfo{volume}{25},
  \bibinfo{pages}{226--250}.
\bibitem[{Verwer and Zhang(2019)}]{verwer2019}
\bibinfo{author}{Verwer, S.}, \bibinfo{author}{Zhang, Y.},
  \bibinfo{year}{2019}.
\newblock \bibinfo{title}{Learning optimal classification trees using a binary
  linear program formulation}, in: \bibinfo{booktitle}{Proceedings of the AAAI
  Conference on Artificial Intelligence}, pp. \bibinfo{pages}{1625--1632}.
\bibitem[{Wang and Lin(2021)}]{Wang2021}
\bibinfo{author}{Wang, T.}, \bibinfo{author}{Lin, Q.}, \bibinfo{year}{2021}.
\newblock \bibinfo{title}{{Hybrid predictive models: when an interpretable
  model collaborates with a black-box model}}.
\newblock \bibinfo{journal}{Journal of Machine Learning Research}
  \bibinfo{volume}{22}, \bibinfo{pages}{1--38}.
\newblock \href{http://arxiv.org/abs/1905.04241}{{\tt arXiv:1905.04241}}.
\bibitem[{Zhou et~al.(2023)Zhou, Athey and Wager}]{zhou2023offline}
\bibinfo{author}{Zhou, Z.}, \bibinfo{author}{Athey, S.},
  \bibinfo{author}{Wager, S.}, \bibinfo{year}{2023}.
\newblock \bibinfo{title}{Offline multi-action policy learning: Generalization
  and optimization}.
\newblock \bibinfo{journal}{Operations Research} \bibinfo{volume}{71},
  \bibinfo{pages}{148--183}.
\bibitem[{Zhu et~al.(2020)Zhu, Murali, Phan, Nguyen and Kalagnanam}]{zhu2020}
\bibinfo{author}{Zhu, H.}, \bibinfo{author}{Murali, P.}, \bibinfo{author}{Phan,
  D.}, \bibinfo{author}{Nguyen, L.}, \bibinfo{author}{Kalagnanam, J.},
  \bibinfo{year}{2020}.
\newblock \bibinfo{title}{A scalable {MIP}-based method for learning optimal
  multivariate decision trees}.
\newblock \bibinfo{journal}{Advances in Neural Information Processing Systems}
  \bibinfo{volume}{33}, \bibinfo{pages}{1771--1781}.

\end{thebibliography}
\endgroup



\end{document}